\documentclass{article}

\usepackage{arxiv}

\usepackage[utf8]{inputenc} 
\usepackage{hyperref}       
\usepackage{booktabs}       
\usepackage{amsfonts}       
\usepackage{nicefrac}       
\usepackage{microtype}      
\usepackage{lipsum}

\usepackage{graphicx}
\usepackage{amssymb}
\usepackage{amsmath}
\newcommand{\argmax}{\mathop{\mathrm{argmax}}}          
\usepackage{subfig}
\usepackage{listings}
\usepackage{times}
\usepackage{latexsym}
\usepackage{url}
\usepackage[ruled,linesnumbered]{algorithm2e}
\usepackage[T3,OT2,T1]{fontenc} 
\usepackage[noenc]{tipa}
\usepackage{tikz}
\usepackage{multirow} 
\usepackage{enumitem} 
\newtheorem{theorem}{Theorem}

\newtheorem{lemma}{Lemma}
\newtheorem{proof}{Proof}
\newtheorem{proposition}{Proposition}
\SetKwRepeat{Do}{do}{while}
\newcommand*{\QEDA}{\null\nobreak\hfill\ensuremath{\blacksquare}}%

\title{Conclusive Local Interpretation Rules for\\ Random Forests}

\author{
  Ioannis Mollas\\
  Aristotle University of \\Thessaloniki, 54636, Greece\\
    \texttt{iamollas@csd.auth.gr}\\
     \And
       Nick Bassiliades\\
  Aristotle University of \\Thessaloniki, 54636, Greece\\
             \texttt{nbassili@csd.auth.gr}\\
             \And
               Grigorios Tsoumakas\\
  Aristotle University of \\Thessaloniki, 54636, Greece\\
             \texttt{greg@csd.auth.gr}\\

}

\begin{document}
\maketitle

\begin{abstract}
In critical situations involving discrimination, gender inequality, economic damage, and even the possibility of casualties, machine learning models must be able to provide clear interpretations for their decisions. Otherwise, their obscure decision-making processes can lead to socioethical issues as they interfere with people's lives. In the aforementioned sectors, random forest algorithms strive, thus their ability to explain themselves is an obvious requirement. In this paper, we present LionForests, which relies on a preliminary work of ours. LionForests is a random forest-specific interpretation technique, which provides rules as explanations. It is applicable from binary classification tasks to multi-class classification and regression tasks, and it is supported by a stable theoretical background. Experimentation, including sensitivity analysis and comparison with state-of-the-art techniques, is also performed to demonstrate the efficacy of our contribution. Finally, we highlight a unique property of LionForests, called conclusiveness, that provides interpretation validity and distinguishes it from previous techniques.


\end{abstract}


\section{Introduction}

It is apparent that machine learning (ML) models will be integrated into our society and daily life. However, in critical domains such as the aforementioned, where models can contain errors, or may suffer from biases, it is more than necessary to ensure their transparency. Gender inequality~\cite{womanCard}, inappropriate patient treatments~\cite{watsonAI} or easily tricked models~\cite{pattern1} are only a few common problems when automated systems are used. Thereby, it is vital for secure, fair, and trustworthy intelligent systems to be able to explain how they work and why they predict a particular outcome. In addition, guaranteeing certain virtues on the behaviour of a model directly allows them to be compliant with legal frameworks, such as the General Data Protection Regulation (GDPR)~\cite{gdpr} of the EU and the Equal Credit Opportunity Act of the US~\footnote{ECOA 15 U.S. Code \S1691 et seq.}. These needs introduced a new area called explainable artificial intelligence (XAI) in the research community~\cite{dovsilovic}. Interpretable machine learning (IML), a subfield of XAI, attempts to address these issues, proposing techniques to shed light into the inner workings of ML models~\cite{adadi,rudin2019stop,bodria2021benchmarking}. 

Random forest (RF) is a highly accurate learning algorithm~\cite{delgado14a} 
that has been proven robust to overfitting~\cite{randomForests}, as well as to learning difficulties arising from phenomena such as class imbalance, or noisy and anomalous data~\cite{VensC11}. RF excels in a lot of sectors. From applications related to security, such as intrusion detection~\cite{intruDe}, to the financial sector dealing with tasks including credit card fraud detection~\cite{fraudDetection} and loan approval~\cite{loanForecast}. The presence of RF algorithms is also vivid in healthcare applications, like the patient safety culture~\cite{SIMSEKLER2020107186} and the classification of different stages of Parkinson's disease~\cite{parkinson}. Finally, in industry and law sectors, applications such as fault diagnosis in self-aligning conveyor idlers~\cite{fdinf} and prediction of crime hotspots~\cite{crimeHotspots} are few examples among many.

The positive aspects of RFs in conjunction with their black box nature have drawn the attention of the IML research community, which introduced a variety of techniques to provide interpretation in the form of rules, trees, or feature importance. For RF models in particular, interpretation techniques follow two main directions. The first one is to create a surrogate model, intending to distil the knowledge of a complex RF model into a single tree~\cite{satoshiTrees,inTrees}. The other direction is to take full advantage of the inner structure of the RF and derive information from the individual trees that make it up~\cite{badCHIRPS,moore,lionForests}. However, these approaches have significant limitations. For example, most of them are only applicable to binary classification tasks~\cite{moore,lionForests}. Another point to consider is that the interpretations produced are not always valid~\cite{badCHIRPS,satoshiTrees}. This is particularly evident in techniques that seek to approximate the actual interpretations of a complex model. 

This work presents LionForests (LF), an approach for interpreting individual predictions of RF models. LF performs path and feature reduction towards smaller interpretation rules with wider ranges, by using unsupervised techniques, such as association rules and $k$-medoids clustering, as well as a path-oriented dissimilarity metric. We extend our preliminary work~\cite{lionForests} in multiple dimensions. 
First of all, we provide a stable theorem and a property, called conclusiveness, to support the validity of the produced rules. Besides that, we broaden LF's capabilities making it applicable to multi-class classification tasks, and we introduce a new concept, the \textit{``allowed error''}, to render the technique applicable to regression tasks as well. In addition, we optimise LF's performance in terms of response time per produced interpretation. Further parameterisation options are also implemented in the technique, concerning the association rules and clustering algorithms. An improved and clearer method for dealing with categorical features is also provided. Moreover, a simple visualisation functionality is designed to increase the expressiveness and clarity of the generated interpretation rules. Experiments including sensitivity analysis, analysis of response time and comparison with other state-of-the-art (SOTA) interpretation techniques, as well as qualitative examples of actual interpretations, are provided to support LF's efficiency. Finally, we are investigating whether SOTA techniques have the same conclusiveness property as LF.


The remainder of this paper is structured as follows. Section 2 presents the related work, while Section 3 establishes the theoretical background. Section 4 describes the LF approach. Exhaustive experimentation from multiple perspectives, such as sensitivity analysis, time analysis, comparison with other methods, and qualitative assessments, is provided in Section 5. Section 6 presents an analysis of the experimental results. Finally, in Section 7 we discuss conclusions and future directions.

\section{Related work}

Over the last few years, IML has advanced so much, making the range of solutions provided wider than ever. Of the several dimensions of interpretability, those primarily used for categorisation and comparison of such approaches refer to the global-local aspect of the model and the agnostic-specific applicability of the technique to a model or architecture. Local-based techniques concern the interpretation of predictions for a single instance, while global-based techniques uncover the entire structure of a model. Both local and global based approaches include techniques that are either model-agnostic or model-specific. Model-agnostic approaches have the potential to interpret any machine learning model indifferently, while model-specific techniques interpret particular types of models, or even architectures. 

Another interesting aspect of interpretation techniques is the form in which they deliver their results. It is possible for a technique to provide an explanation as a set of rules, a set of feature weights, images with skewed or highlighted sections, or prototype examples. In this section, we will present a few techniques related to these dimensions, in the context of ensemble models, such as random forests. These techniques share the same rule-based interpretation form. This means that the output of these techniques is a set of rules, or single rules, explaining a model or a specific prediction.

Surrogate models~\cite{surrogateModels}, based on the principle of model compression~\cite{modelCompression}, are a global-based, model-agnostic interpretation technique that attempts to imitate the behaviour of more complex models. A decision tree (DT), for example, is a surrogate model of an RF model, when trained on training data, labelled by the RF model. Metrics such as fidelity emerged in order to evaluate the ability of these models to mimic the original models. High fidelity means the approximation is sufficient. However, it is still questionable how well the surrogate models should approximate black-box models in order to be trusted. 

Anchors~\cite{anchors} and LORE~\cite{italianDude}, local-based model-agnostic techniques leveraging sub-spaces of the original data, create local surrogate models to explain particular instance predictions. Specifically, Anchors provides a single rule for interpreting the prediction of a single instance. This rule is the anchor that keeps the prediction (almost, as the authors acknowledge) always the same. Anchors is only applicable to datasets with discrete features. On the other hand, for each instance, LORE creates a local neighbourhood using genetic algorithms to construct a surrogate decision tree from which a single rule interpretation and counterfactual instances are extracted.

The single-tree approximation approaches like inTrees~\cite{inTrees} and defragTrees~\cite{satoshiTrees} among others~\cite{domingos,zhouTrees}, are global model-specific techniques that interpret tree ensembles by approximating the output of the model they attempt to explain through a single tree. However, this procedure is highly problematic because it is not feasible to summarise a complex model like tree ensembles to a single simple tree, as reported by other researchers~\cite{donotusesingletreesappx}. 

\begin{figure}[ht]
\centerline{\includegraphics[width=0.9\textwidth]{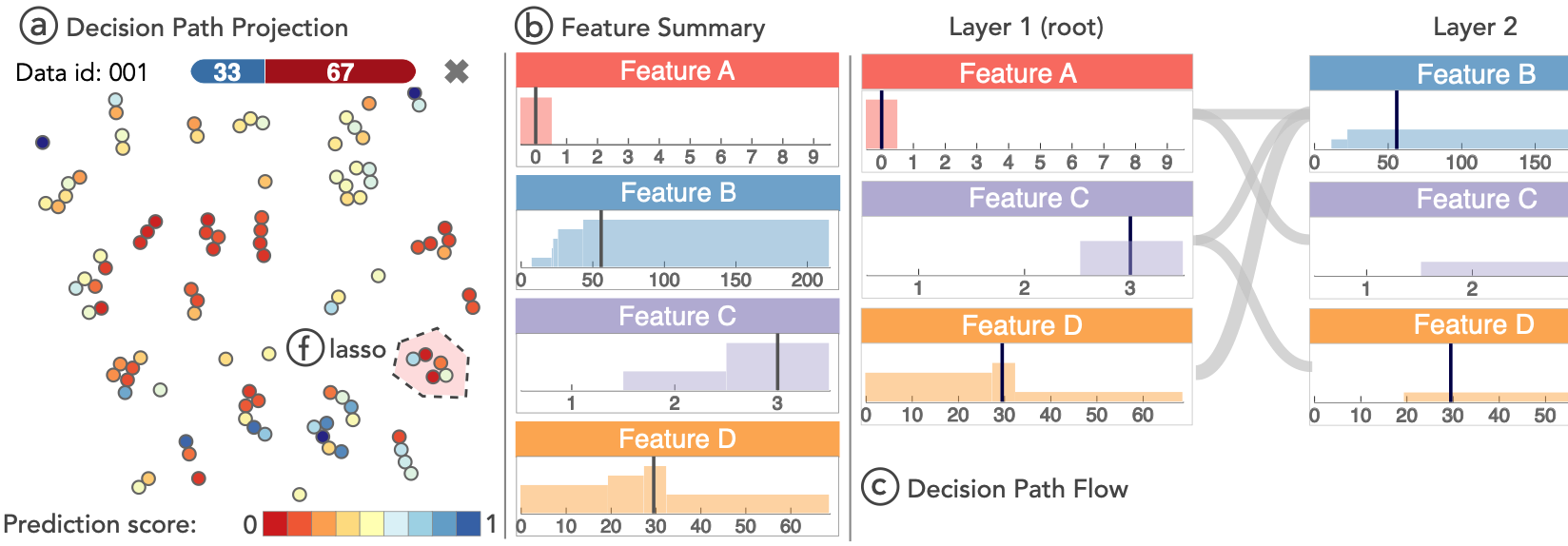}}
\caption{A sample example from the visualisation tool of iForest adapted from~\cite{iforest}} \label{fig:iForest}
\end{figure}

iForest~\cite{iforest} is a visualisation method that offers global and local interpretations of RF models. The most interesting aspect of this tool, though, concerns local explanations. Using a path distance metric, iForest projects the paths of an instance to a two-dimensional space (Figure~\ref{fig:iForest}a) using t-Distributed Stochastic Neighbour Embedding (t-SNE)~\cite{tsne}, and it provides a feature summary (Figure~\ref{fig:iForest}b) and a decision path flow (Figure~\ref{fig:iForest}c), as well. iForest requires user input, in the form of drawing a lasso around a set of paths (Figure~\ref{fig:iForest}f), to produce the path flow. This is a disadvantage, because the user may give wrong input, leading to an incorrect feature summary and path flow, which will result in a flawed interpretation. 

\begin{figure}[ht]
\centering
\minipage{0.395\textwidth}
\centerline{\includegraphics[width=\linewidth]{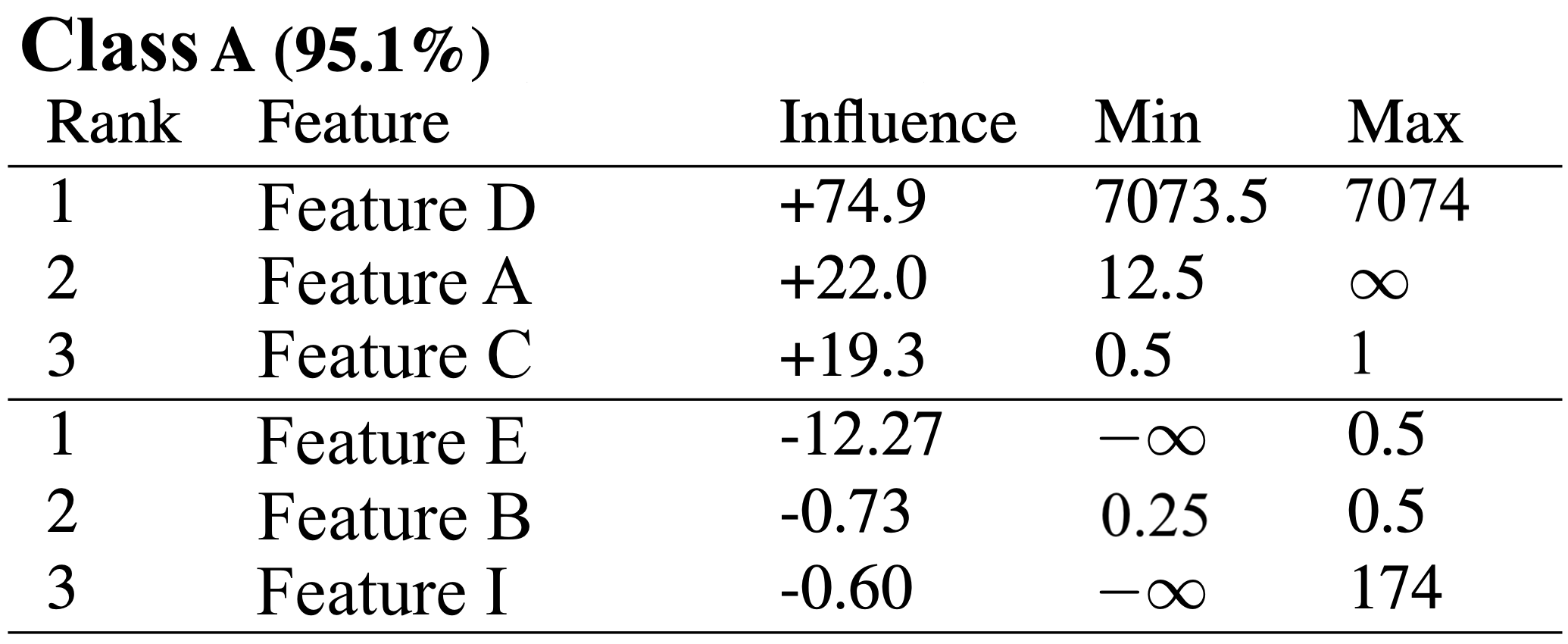}}
\caption{Template of explanation of Moore et al.~\cite{moore}} \label{fig:moore}
\endminipage\hfill
\minipage{0.58\textwidth}
\centerline{\includegraphics[width=\linewidth]{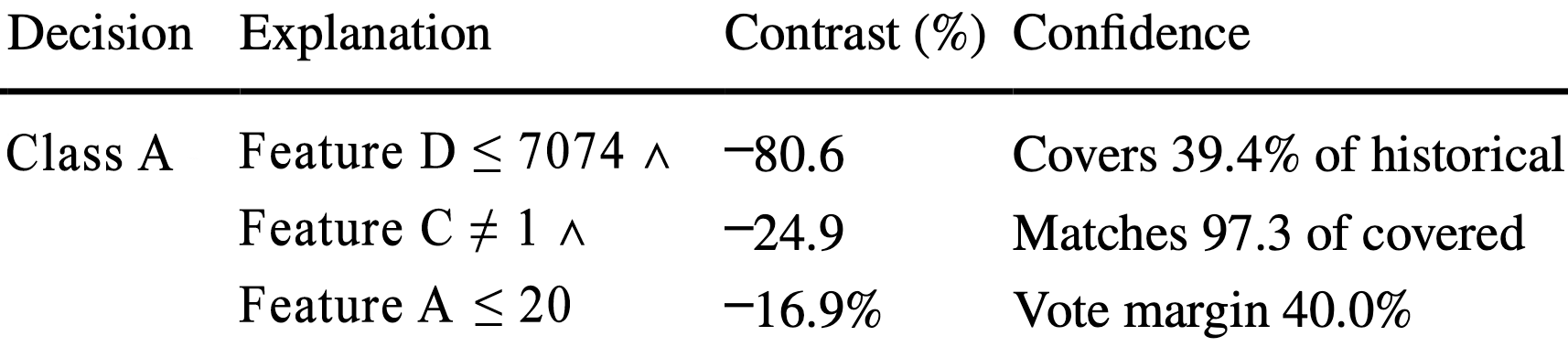}}
\caption{Template of explanation of CHIRPS~\cite{badCHIRPS}} \label{fig:chirps}
\endminipage
\end{figure}

Another local model-specific interpretation strategy~\cite{moore} interprets RF models by providing a collection of features with their ranges, rated on the basis of their importance, as explanations (Figure~\ref{fig:moore}). The process of interpretation consists of two steps. First, they measure the effect of a feature $j$ on the prediction for a given instance $x$, and later this effect will be used for the ranking process. To accomplish this, a node per tree monitoring mechanism is used to find the aggregated effect of all the features for the prediction of a particular instance. The second step is to identify the narrowest range for each feature over all trees.

Finally, CHIRPS~\cite{badCHIRPS} proposes a technique for multi-class tasks, using frequent pattern (FP) mining on the paths of the majority class to identify the most influencing features and present them to the end user. CHIRPS promises interpretations (Figure~\ref{fig:chirps}) that will be minimally complete, providing information about counterfactual cases, and referring to real data, and not synthetic data. A drawback of this approach is that it lacks a strict restriction about the number of paths that will be covered through the generated rule, particularly in multi-class classification tasks. 
An extension of CHIRPS to gradient boosted tree ensembles is gbt-HIPS~\cite{gbtHips}.


LF overcomes the issues of single-tree approximations, does not require user input like iForest, and comes with low computational cost in contrast to Anchors. Most importantly, LF provides rules that are always valid. Finally, it is applicable in a wider range of machine learning tasks compared to Anchors, LORE and CHIRPS, among other competitors.

\section{Main concepts and notation}

We here define the main concepts and notation concerning decision trees and random forests, which are necessary for the presentation of LF in the next section.

\subsection{Decision trees}

\begin{figure}[ht]
    \centerline{\includegraphics[height=2in]{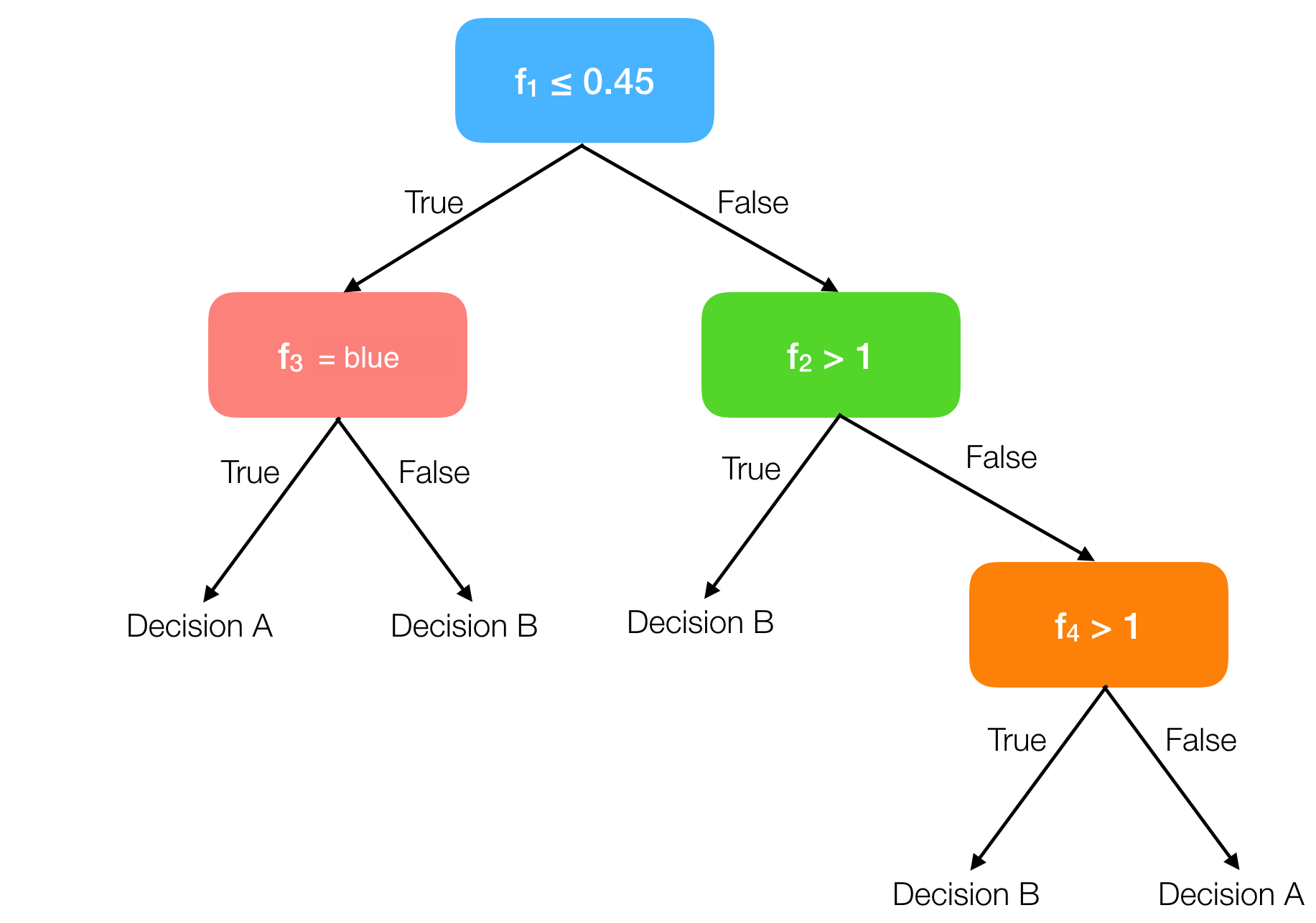}}
    \caption{A simple decision tree classifier with 4 features} 
    \label{exampledt}
\end{figure}

Decision Trees (DT)~\cite{DTrees} is a classic machine learning algorithm that stimulates the growth of the Ensemble algorithms. A DT can be shown as an acyclic directed graph, as seen in Figure~\ref{exampledt}, containing a root node, decision nodes and leaf nodes, which are the prediction nodes. Regarding the learning algorithm, each node concerns a particular feature $f_i$ and a condition relation. In the case of input instances, the decision tree traces the path to a leaf node containing a prediction. The prediction can be a class, in the case of a binary or multi-class classification, or a real number estimate, in the scenario of a regression task.

Each decision path $p$ is a conjunction of conditions, and the conditions are features and values with relations $\leq$, $>$ and $=$. To provide an illustration, a path from the tree on Figure~\ref{exampledt} would be: `if $f_1 > 0.45$ and $f_2 \leq 1$ and $f_3 = blue$ then \textit{Decision A}'. Thus, each path $p$ is expressed as a set:\begin{equation} p=\{f_i \boxtimes v_j | f_i \in F, ( ( \boxtimes \in \{\leq, >\} \rightarrow v_j \in \mathbb{R}) \wedge (\boxtimes \in \{=\} \rightarrow v_j \in S_j ) ) \}\end{equation}


where $v_j$ the instance's value for the feature $f_i$, and $S_j$ is the set of the categorical values if $f_i$ is a discrete feature.

However, LionForests implementation relies on a library\footnote{We use scikit-learn as core library (\url{https://scikit-learn.org})}, which uses an optimised version of DTs, where the categorical features must be either encoded to numerical features, with encoding procedures like OneHot or Ordinal encoding. Hence, in the rest of the paper each path $p$ is expressed as a set:\begin{equation} p=\{f_i \boxtimes v_j | f_i \in F, v_j \in \mathbb{R}, \boxtimes \in \{\leq, >\}\}.\end{equation}

\subsection{Random forests}
One of the first ensemble algorithms using DTs as foundation is the Random Forests (RF)~\cite{randomForests} algorithm. RF is a collection of a specific number of trees, which are combined under an equal voting scheme. Abstractly, the inference process of an RF model can be seen as a voting procedure where the outcome (prediction) corresponds to the majority of the votes cast.

These trees are trained under different data and feature partitions, towards higher variance and lower bias, dealing with the overfitting problem. Then, for a specific prediction the trees are voting: \begin{equation} \label{eq:1} h(x_i) = \dfrac{1}{|T|} \sum_{t \in T}^{} h_t(x_i) \end{equation} where $h_t(x_i)$ is the vote cast from the tree $t \in T$ for the instance $x_i \in X$, representing the probability $P(C=c_j|X=x_i)$ of $x_i$ to be assigned to class $c_j \in C$, and finally choosing the $\argmax_{c \in \mathcal{C}} P(C=c|X=x_i)$ for classification problems, and $h_t(x_i) \in \mathbb{R}$ in regression tasks.




\section{Our approach}

\begin{figure}[ht]
    \centerline{\includegraphics[width=1\textwidth]{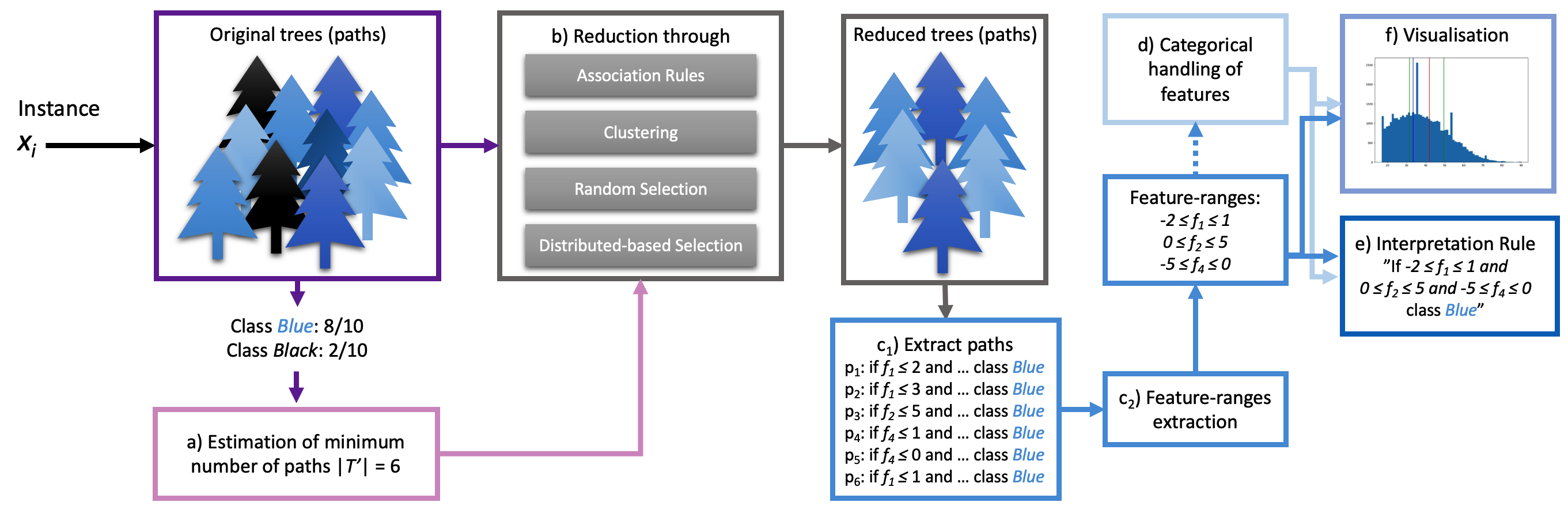}}
    \caption{LionForests architecture} 
    \label{lfarch}
\end{figure}

We present the core of LionForests (LF), along with its extensions, which concern primarily the theoretical grounding, the optimisation of the main algorithm, as well as the adaptation from binary classification to multi-class and regression tasks. 

LF is a technique for local interpretation of binary, multi-class and regression RF models. This is achieved via the following sequence of actions: a) estimation of minimum number of paths, b) reduction through association rules, clustering, random selection, or distribution-based selection, c) extraction of feature-ranges, d) categorical handling of features and, ultimately, e) composition of interpretation and f) visualisation, as depicted in Figure~\ref{lfarch}. Before presenting to these actions, we introduce a new property called {\em conclusiveness} (in the following subsection). In order to produce rules with fewer features and broader ranges, LF first estimates the minimum number of paths that has to be maintained to produce {\em conclusive} rules. Then, it reduces the redundant paths and extracts the feature-ranges from the rest. In case of categorical features, a related procedure takes place. Finally, a single rule in natural language is provided, as well as a visualisation of the rule's details.

\subsection{Conclusiveness property}\label{sec:concl}

\begin{quote}
    \textit{conclusiveness - the quality of being final or definitely settled}
\end{quote}

We here introduce a property which concerns an aspect of a rule's quality, which we call {\em conclusiveness}. Thus, in a scenario with a dataset with e.g. 4 features ($F = [f_1, f_2,$ $f_3, f_4])$, a predictive model $h$, an instance $x$ with values for the corresponding features $[v_1,v_2,v_3,v_4]$, and the prediction $c_k$, $h(x)=c_k$,, we have a rule ``$r =$ if $2\leq f_1 \leq 10$ and $0.5\leq f_3 \leq 1.2$ then $c_k$'' containing conditions with 2 of them ($F'=[f_1, f_3]$). \begin{equation} r=\{f_i \boxtimes v_j \rightarrow c_k | f_i \in F, v_j \in \mathbb{R}, \boxtimes \in \{\leq, >\}\}.\end{equation}

\[
  \textrm{conclusive}(r,c_k) = 
  \begin{cases}
    False, & \exists v_l \in D(f_i), f_i = v_l\boxtimes v_j | f_i\boxtimes v_j \in r, h(x'_{f_i=v_l}) \neq c_k \\
    False, & \exists v_l \in D(f_i), f_i = v_l| f_i \notin r, h(x'_{f_i=v_l}) \neq c_k \\
    True, & \textrm{elsewhere}
  \end{cases}
\]

,where $D(f_i)$ is the domain of feature $f_i$, and $x'_{f_i=v_l}$ is an alteration of instance $x$ but with different value ($v_l$) for a specific feature $f_i$. Therefore, a rule will be conclusive if and only if the following restrictions are met: 

\begin{enumerate}
    \item if the instance's values for the excluded features are modified to \underline{\textit{any possible}} \underline{\textit{value}} the prediction will not be influenced,
    \item if the value of one of the included features is modified \underline{\textit{within the specified range}} the prediction will not be influenced.
\end{enumerate}

A technique that always produces conclusive rules owns the property of conclusiveness. As presented in the experiments of Section~\ref{sec:exp}, however, a conclusive rule contains more features than an inconclusive one, increasing the rule length, and is specific to a particular instance, reducing the rule's coverage.

\subsection{Estimation of minimum number of needed paths}\label{mnpr}

LF attempts to identify and use only a subset of the paths that voted for the predicted class ($C_{M}$), in order to reduce the features appearing in a rule and to broaden the ranges of the features. 
However, to acquire the conclusiveness property, LF always has as many paths as needed to remain consistent with the original prediction. For each learning task (binary, multi-class and regression) it is necessary to define the theoretical foundations on which the estimation of the minimum number of needed paths will rely on.

We should mention here that in the implementation of LF, apart from the minimum number of paths, an additional restriction concerning the average probability of a class is added. This happens because unlike the original RF's majority voting schema (hard voting)~\cite{randomForests}, the implementation of RF we use \cite{sklearn} combines the probabilistic predictions (soft voting) of the trees.

\subsubsection{Minimum number of paths in binary tasks}
\label{estb}
First, we will go through the theoretical foundations for estimating the minimum required paths on binary tasks. Based on Proposition~\ref{proposition1}, which states that we need at least $\frac{|T|}{2}+1$ of the paths in order to maintain the same prediction $C_{M}$, LF will select at least a quorum. For example, if we had $|T| = 100$ trees in an RF model and $89$ of them voted for class M for an instance $x_i$, it has to select at least $51$ of the $89$ paths extracted from the trees, in order to produce the interpretation rule.

\begin{proposition}
\label{proposition1}
An RF model, with $T$ trees casting $|T|$ votes, predicts always class M ($C_{M}$) if and only if class M has at least a $quorum$ of votes or more, where $quorum$ $= \frac{|T|}{2}+1$ out of $|T|$ votes.
\end{proposition}

As a result, the optimisation problem (Eq.~\ref{eq:optimazation}) is formulated to minimise the number of features ($|F'|$) that satisfy a reduced set of paths ($|T'|$). The constraint is to retain the same classification result as the original set of trees, making the number of the reduced paths equal to or greater than the quorum. Furthermore, the final interpretation rule will include any feature that appears in these paths, guaranteeing the property of conclusiveness.

\begin{equation}
\label{eq:optimazation}
\begin{aligned}
& \underset{F' \subseteq F}{\text{minimise}}
& & |F'|\\
& \text{subject to}
& & p=\{f_i \boxtimes v_j | f_i \in F'\},  p \in P_t \text{ } \forall \text{ } t \in T', \\
& \text{     }
& & \lfloor \frac{1}{|T|} \sum_{t \in T'}^{} h_t(x_i) + \frac{1}{2} \rfloor = \lfloor\frac{1}{|T|} \sum_{t \in T}^{} h_t(x_i) + \frac{1}{2}\rfloor,\\
& \text{     }
& & |T'| \geq \frac{|T|}{2}+1
\end{aligned}
\end{equation}

An example for the equation $\lfloor \frac{1}{|T|} \sum_{t \in T}^{} h_t(x_i) + \frac{1}{2} \rfloor$ follows. When $70$ out of $|T|=100$ trees are voting for class 1, then we have $\lfloor\frac{1}{100}70 + 0.5\rfloor=\lfloor1.2\rfloor \rightarrow 1$. On the other hand, if 25 out of $|T|=100$ trees are voting class 1 (the minority), then we have $\lfloor\frac{1}{100}25 + 0.5\rfloor=\lfloor0.75\rfloor \rightarrow 0$. Therefore, we are aiming to find the smallest $T'\subseteq T$, which will produce the same classification as the original $T$ trees.

\subsubsection{Minimum number of paths in multi-class tasks}
\label{estm}
In a multi-class classification task, the voting system is more complex than in a binary task, and thus Proposition~\ref{proposition1} cannot cover every case. The following theorems are presented in order to apply the framework of LF to multi-class tasks. A running example will be given, along with the propositions and theorems, for better comprehension. 

An RF model always predicts class M ($C_M$) if and only if class M has a majority of votes, while any other class J ($C_J$) has fewer votes. Let, the available classes for the RF model with $|T| = 100$ trees be $C_1$ (\textit{red}), $C_2$ (\textit{blue}) and $C_3$ (\textit{green}). For a specific case, each tree voted and the result was $|C_1|$ $=45$, $|C_2|=35$ and $|C_3|=20$. Thus, the prediction was \textit{red} with $|C_{1}| = |C_M| =45$, and the second highest voting class was \textit{blue} with $|C_{2}|=35$. Proposition~\ref{proposition1} cannot be used in this example because $45=|C_{1}|=|C_{M}|<quorum=\frac{|T|}{2}+1=51$.

Based on this, the result of the RF model is class M with the majority of votes. However, if we minimise the voting paths to the majority class votes, as we did in binary tasks (only if the number of votes is less than the quorum), and extract the interpretation rule, it is possible for another class to gain votes from other classes, because the paths voting for those classes will not be covered by the rule, and thus their vote (prediction) can change to the various paths of the decision trees that might arise, and exceed the votes of the originally first class. As a consequence, the outcome of the prediction could change.

\begin{lemma}
\label{lemma1}
An RF model always predicts class M ($C_M$) if and only if class M has a majority of votes, while any other class J ($C_J$) can not exceed the votes of class M by obtaining at least $S = |C_{M}| - |C_J| + 1$ votes from the other classes.
\end{lemma}

In our running example, if we apply Lemma~\ref{lemma1}, we can assume that RF will always predict the \textit{red} class if the other classes do not exchange votes. For example, if $|C_{1}| - |C_{2}| + 1 = 45 - 35 + 1 = 11$ votes from the \textit{green} class move to the \textit{blue} class, the number of votes from the \textit{blue} class would rise to $|C_2|=35 + 11 = 46$, which are more than the \textit{red} class. Or if $|C_{1}| - |C_{3}| + 1 = 45 - 20 + 1 = 26$ votes from the \textit{blue} class move to the \textit{green} class, the number of votes from the \textit{green} class would rise to $|C_3|=20 + 26 = 46$ more than the \textit{red} class. As a result, if we had reduced the paths to the number of class \textit{red} votes with LF, then the other votes could change and, therefore, the outcome of the prediction could also change, while Lemma~\ref{lemma1} would be invalid.

\begin{proposition}
\label{proposition2}
An RF model always predicts class M ($C_{M}$) if and only if the $K$ number of votes from any other class remains stable, where $K = |T| + |C_{L}| - |C_{M}| + 1$, including the $C_{M}$ and $C_{L}$, votes from the majority class and the second most voted class, respectively, out of the $|T|$ votes.
\end{proposition}

Proposition~\ref{proposition2} argues that in situations where the majority of the class has less than a quorum of votes, the number of votes can be reduced to $K$ without affecting the outcome of the prediction. Maintaining only the votes of the two key classes, majority class and second most voted class, is not enough and hence we need to retain $K_{other} = K - |C_{M}| - |C_{L}|$ from the other classes, also at random.

Thus, in our running example, the minimum number of paths we can reduce to is $K = 100 + 35 - 45 + 1 = 91$. From these 91 paths, we will retain the 45 of class `red' ($C_{1}$) and the 35 of class `blue' ($C_{2}$) paths. Then we would have to hold $K_{other} = 91 - 45 - 35 = 11$ paths from the remaining classes, in this example from the `green' class ($C_{3}$) which holds 20 paths. We use the LF's techniques, which will be detailed later in this section, to select 11 out of 20 paths. Thus, holding 91 paths out of 100, the rest of the 9 paths of the class `green', if they all switch from `green' to `blue', the class `blue' will not exceed the votes of the class `red', and the result will remain the same. This is presented in Theorem~\ref{def4}.

\begin{theorem}
\label{def4}
For $R = |T| - |C_{M}| - |C_{L}|$ as the remainder of the votes and $K_{other} = K - |C_{M}| - |C_{L}|$ as the remainder of the votes to be selected, $S$ is always greater than $R - K_{other}$, which means that there is not a sufficient number of votes to be received by class $L$, or any other class, in order to surpass the votes of class $M$.
\end{theorem}

In order to prove that the aforementioned Theorem holds, we will use contradiction to prove that $R - K_{other} < S$ is true.

\begin{proof}
\label{proof}
To prove Theorem~\ref{def4} by contradiction we assume that the statement $R - K_{other} < S$ is false. We will prove that $R - K_{other} \geq S$ is true.

\begin{align*}
    R - K_{other} \geq S  & \Longleftrightarrow \\
    (|T| - |C_M| - |C_L|) - (K - |C_M| - |C_L|) \geq |C_M| - |C_L| + 1 & \Longleftrightarrow \\
    |T| - |C_M| - |C_L| - K + |C_M| + |C_L| \geq |C_M| - |C_L| + 1 & \Longleftrightarrow \\
    |T| - K \geq |C_M| - |C_L| + 1 & \Longleftrightarrow \\
    |T| - (|T| + |C_{L}| - |C_{M}| + 1) \geq |C_M| - |C_L| + 1 & \Longleftrightarrow \\
    |T| - |T| - |C_{L}| + |C_{M}| - 1 \geq |C_M| - |C_L| + 1 & \Longleftrightarrow \\
    |C_{M}| - |C_{L}| - 1 \geq |C_M| - |C_L| + 1 & \Longleftrightarrow \\
    - 1 \geq 1 & 
\end{align*}

Thus, we proved that the statement $R - K_{other} \geq S$ is not true. \QEDA
\end{proof}

Therefore, the number of paths we can keep in order to always maintain the same classification result, is provided by the following rules:
\begin{itemize}
    \item If $|C_{M}| \geq quorum$ then apply LF reduction to features and paths as it happens to binary tasks, based on Proposition~\ref{proposition1},
    \item If $|C_{M}| < quorum$ then identify the class with the second highest amount of votes, denoted as $C_{L}$. The number of paths we need to keep is equal to $quorum'=|C_{L}| - |C_{M}| + T + 1$. Then, we keep the paths from the $C_{M}$ and $C_{L}$, and a random selection process is employed to collect $quorum'$ from the $R$ remaining paths.
\end{itemize}


\subsubsection{Minimum number of paths in regression tasks}
\label{estr}
The estimation of the required number of paths as defined in the binary and multi-class tasks, are not applicable for regression. Hence, we introduce an algorithm to adapt LF to regression models. In RF for regression, the prediction is determined by the average of the individual trees' predictions of the RF,
\begin{equation} \label{eq:3} h(x_i) = \dfrac{1}{|T|} \sum_{t \in T}^{} h_t(x_i) \end{equation}
where $h_t(x_i) \in \mathbb{R}$. On this basis, we cannot assume that by removing a few trees (for example, preserving a quorum) we can get the same outcome. To overcome this, we introduce the $allowed\_error$ definition.

The idea is to reduce the number of features and paths with regard to an $allowed\_$ $error$. We set the mean absolute error of the RF model as the default value for $allowed\_error$. We also let the user set a preferred $allowed\_error$. A sensitivity analysis between $allowed\_error$ and the feature and path reduction ratio is provided in Section~\ref{sec:sensanalysis} to guide users to choose the appropriate $allowed\_error$ for their application based on their needs. The value of $allowed\_error$ is strongly associated with the importance of each task. Saying we have a regression model that predicts the indication of blood sugar. The importance of the error has a mandatory sense, since it involves a health issue. On the other hand, when forecasting e.g. the quality of wine, the error might be less sensitive.


In order to measure the error of an interpretation and to use the following reduction techniques, the min and max predictions per tree of RF must be collected. Thus, for each tree of an RF model, we traverse to its leaves to identify the min and max predictions that the tree can provide. This kind of information will be used to estimate the worst-case error that may be added to the prediction given an interpretation. Therefore, the final interpretation will have the following form:

\begin{quote}
\centering
    \smaller{if $0.47 \leq f_1 \leq 0.6$ and $22 \leq f_3 \leq 54$ then Prediction: $24.15\pm local\_error$, \\where $local\_error\in [0,allowed\_error]$}
\end{quote}

\begin{algorithm}[ht]
 \KwIn{$trees, allowed\_error, tree\_stats$}
 \KwOut{$local\_error$}
    $L, K \gets reduction\_method(trees, allowed\_error)$\\
    $original\_pred \gets 0$, $prediction \gets 0$ \\
    \For{$tree \in L$}{
        $original\_pred \gets original\_pred + tree.predict(instance)$\\
        $prediction \gets prediction + tree.predict(instance)$\\
    }
   \For{$tree \in K$}{
        $tree\_pred = tree.predict(instance)$\\
        $min\_pred = tree\_stats[tree].min$\\
        $max\_pred = tree\_stats[tree].max$\\
        $extreme\_pred \gets max\_pred$ \\
        \If{$|tree\_pred - min\_pred| > |tree\_pred - max\_pred|$}{
            $extreme\_pred \gets min\_pred$ \\
        }
        $original\_pred \gets original\_pred + tree\_pred$\\
        $prediction \gets prediction + extreme\_pred$\\
    }
    $prediction = \frac{prediction}{|L+K|}$\\
    $original\_pred = \frac{original\_pred}{|L+K|}$\\
    $local\_error \gets |original\_pred - prediction|$\\
    \KwRet{$local\_error$}
 \caption{Calculation process of $local\_error$}
 \label{alg:localerror}
\end{algorithm}

The $local\_error$ is determined with regard to a selection of paths $|L|$. Three algorithms are introduced, presented in the following paragraphs, that will lead to a selection of $|L|$ trees from the initial $|T|$ trees. The $|K|$ remaining trees, which will be excluded from the final interpretation rule, can cause an error in the prediction. That is because the rule does not completely cover the decision of these trees, and thus, for a change in the instance to a feature that does not appear in the rule, those trees may produce different predictions, and the final prediction may change to:

\begin{equation} \label{eq:4} 
h'(x) = \dfrac{1}{|T|} (\sum_{t \in T-K}^{} h_t(x_i) + \sum_{t \in K}^{}e_t(x_i))
\end{equation}
\[
e_t(x_i) = 
\begin{cases}
    min\_pred_t, & |h_t(x_i) - min\_pred_t| > |h_t(x_i) - max\_pred_t| \\
    max\_pred_t, & elsewhere
\end{cases}
\]

Consequently, for those $|K|$ trees, we replace their predictions with the min or max prediction that each tree can provide, choosing between them on the basis of which, min or max, is the most distant to the original prediction of the tree. The aforementioned process is depicted in the Algorithm~\ref{alg:localerror}, which is the base for Algorithm~\ref{alg:dsred} presented later in Section~\ref{sec:reddbs}.

\subsection{Reduction techniques}\label{redTechs}
We now introduce the reduction techniques applied after identifying how many paths we can reduce. We define four reduction techniques that we use to build rules with fewer features and larger ranges. Table~\ref{red:algorithms} depicts an overview of the applicability of each algorithm presented in the following sections to the corresponding tasks. All of the techniques aim to reduce both the paths and the features, but with different degrees of effectiveness.

\begin{table}[ht]
\centering
\resizebox{\textwidth}{!}{%
\begin{tabular}{lrccccc}
\cline{3-7}
 &
  \multicolumn{1}{l|}{\multirow{3}{*}{}} &
  \multicolumn{5}{c|}{Reduction through} \\ \cline{3-7} 
 &
  \multicolumn{1}{l|}{} &
  \multicolumn{1}{c|}{\multirow{2}{*}{\begin{tabular}[c]{@{}c@{}}Association\\ Rules\end{tabular}}} &
  \multicolumn{1}{c|}{\multirow{2}{*}{Clustering}} &
  \multicolumn{1}{c|}{\multirow{2}{*}{\begin{tabular}[c]{@{}c@{}}Random\\ Selection\end{tabular}}} &
  \multicolumn{2}{c|}{\begin{tabular}[c]{@{}c@{}}Distribution-based\\ Selection\end{tabular}} \\ \cline{6-7} 
 &
  \multicolumn{1}{l|}{} &
  \multicolumn{1}{c|}{} &
  \multicolumn{1}{c|}{} &
  \multicolumn{1}{c|}{} &
  \multicolumn{1}{c|}{Inner} &
  \multicolumn{1}{c|}{Outer} \\ \cline{2-7} 
\multicolumn{1}{l|}{} &
  \multicolumn{1}{r|}{Binary} &
  \multicolumn{1}{c|}{\checkmark} &
  \multicolumn{1}{c|}{\checkmark} &
  \multicolumn{1}{c|}{\checkmark} &
  \multicolumn{1}{c|}{-} &
  \multicolumn{1}{c|}{-} \\ \cline{2-7} 
\multicolumn{1}{l|}{} &
  \multicolumn{1}{r|}{Multi-class} &
  \multicolumn{1}{c|}{\checkmark} &
  \multicolumn{1}{c|}{\checkmark} &
  \multicolumn{1}{c|}{\checkmark} &
  \multicolumn{1}{c|}{-} &
  \multicolumn{1}{c|}{-} \\ \cline{2-7} 
\multicolumn{1}{l|}{} &
  \multicolumn{1}{r|}{Regression} &
  \multicolumn{1}{c|}{\checkmark$_{variant}$} &
  \multicolumn{1}{c|}{-} &
  \multicolumn{1}{c|}{\checkmark} &
  \multicolumn{1}{c|}{\checkmark} &
  \multicolumn{1}{c|}{\checkmark} \\ \cline{2-7} 
 &
  \multicolumn{1}{l}{} &
  \multicolumn{1}{l}{} &
  \multicolumn{1}{l}{} &
  \multicolumn{1}{l}{} &
  \multicolumn{1}{l}{} &
  \multicolumn{1}{l}{} \\ \hline
\multicolumn{1}{|r|}{\multirow{2}{*}{\begin{tabular}[c]{@{}r@{}}Designed\\ Towards\\\end{tabular}}} &
  \multicolumn{1}{r|}{Feature Reduction} &
  \multicolumn{1}{c|}{\checkmark} &
  \multicolumn{1}{c|}{\checkmark} &
  \multicolumn{1}{c|}{-} &
  \multicolumn{1}{c|}{-} &
  \multicolumn{1}{c|}{-} \\ \cline{2-7} 
\multicolumn{1}{|r|}{} &
  \multicolumn{1}{r|}{Path Redution} &
  \multicolumn{1}{c|}{-} &
  \multicolumn{1}{c|}{-} &
  \multicolumn{1}{c|}{\checkmark} &
  \multicolumn{1}{c|}{\checkmark} &
  \multicolumn{1}{c|}{\checkmark} \\ \hline
\end{tabular}%
}
\caption{Overview of algorithms we use for feature and path reduction for each task}
\label{red:algorithms}
\end{table}

\subsubsection{Reduction through association rules}
\label{subsec:rar}

The first reduction process, reduction through association rules (AR), is applicable to all learning tasks with small variations, and it starts with the implementation of the association rules~\cite{associationrules}. In association rules, the attributes are called items $I=\{i_{1},i_{2},\ldots ,i_{n}\}$. Each dataset contains sets of items, called itemsets $T=\{t_{1},t_{2},\ldots ,$ $t_{m}\}$, where $t_i \subseteq I$. Using all possible items of a dataset, we can find all the rules $X\Rightarrow Y$, where $X,Y\subseteq I$. $X$ is called antecedent, while $Y$ is called consequent. The purpose of the association rules is to determine the support and confidence of every rule in order to find useful relations. A straightforward observation is that $X$ is independent of $Y$ when the confidence level is particularly low. What is more we can tell that $X$ with high support implies it is probably very significant.

\begin{algorithm}[ht]
 \KwIn{$paths, features, trees, minimum\_number\_paths/allowed\_error, task$}
 \KwOut{$reduced\_paths, reduced\_feature\_set$}
    $itemsets \gets []$ \tcp*{Empty list of itemsets}
    \For{$path \in paths$}{
        $itemset \gets []$ \\
        \For{$f_i \in path$}{
            $itemset \gets itemset + f_i$
        }
        $itemsets \gets itemsets + itemset$ \\
    }

    $frequent\_itemsets \gets ar(itemsets)$ \tcp*{$ar$ can be apriori, fpgrowth} 
    
    $antecedent\_features \gets []$, $antecedents \gets \emptyset$ \\
    \For{$antecedent \in frequent\_itemsets[antecedents]$}{
        \If{$antecedent \notin antecedents$}{
            $antecedents \gets antecedents + antecedent$\\
            \For{$feature \in antecedent$}{
                \lIf{$feature \notin antecedent\_features$}{
                    $antecedent\_features \gets antecedent\_features + feature$
                }
            }
        }
    }
    $reduced\_paths \gets []$, $reduced\_feature\_set \gets []$\\
    \lIf{$task\in \{regression\}$}{$local\_error \gets 2* allowed\_error$}

    $k \gets 1$ \\
    \While{$((task \in \{binary, multi-class\} \wedge |reduced\_paths| < minimum\_number\_paths) \vee (task \in \{regression\} \wedge local\_error > allowed\_error)) \wedge k < |antecedent\_features|$}{
        $reduced\_feature\_set \gets reduced\_features + antecedent\_features[k-1]$\\
        $redundant\_features \gets []$\\
        \For{$feature \in features$}{
            \lIf{$feature \notin reduced\_feature\_set$}{$redundant\_features \gets redundant\_features + feature$}
        }
        $reduced\_paths \gets []$\\
        \For{$path in paths$}{
            \lIf{$\forall feature \in path, feature \notin redundant\_features$}{
                $reduced\_paths\gets reduced\_paths + path$
            }
        }
        \lIf{$task \in \{regression\}$}{$local\_error \gets compute\_error(paths,reduced\_paths)$}
        $k \gets k + 1$
    }
    \KwRet{$reduced\_paths,reduced\_feature\_set$}
 \caption{Reduction through association rules}
 \label{alg:arred}
 \end{algorithm}

Algorithm~\ref{alg:arred} describes the reduction through association rules. The association rules will be implemented at the path level. The $I$ items will contain the $F$ features of our original dataset. The $T$ dataset, which will be used to mine the association rules, will contain a collection of features that reflect each path $t_{i}=\{i_{j}|i_{j}=f_{j} , f_j \boxtimes v_k \in p_{i}, p_{i}\in P\}$. It is important to mention that we keep only the presence of a feature in the path, and we discard the value of $v_j$, in order to have itemsets only with features. It is then possible to apply association rules methods. To compute the association rules we use apriori~\cite{apriori} and fpgrowth~\cite{fpgrowth}, influenced by comparative studies of association rules algorithms~\cite{arcomp1,arcomp2}, presenting these as options for the reduction through association rules process. 

The next step is to sort the association rules produced by the algorithm based on the ascending confidence score. For the rule $X \Rightarrow Y$ with the lowest confidence, we will take items of $X$ and add them to an empty list of features $F'$. After that, we determine the number of paths containing conjunctions that are satisfied with the new feature set $F'$. We stop when we acquire a number of paths either equal or more than the estimated required paths (as presented in Sections~\ref{estb} and~\ref{estm}), or when we have a $local\_error$ from the paths that cannot be valid with the new feature set, smaller or equal to the $allowed\_error$ (Section~\ref{estr}). 

Otherwise, we iterate and add more features taken from the next antecedent of the next in confidence score rule. By using this strategy, we keep the number of features low and smaller than the original $F' \subseteq F$. Reducing the features can also lead to a reduced collection of paths, as paths containing conjunctions with redundant features would no longer be valid. So, we have the following representation for every $p$ path: \begin{equation}p=\{f_i \boxtimes v_j | f_i \in F', v_j \in \mathbb{R}, \boxtimes \in \{\leq, >\}\}.\end{equation}

It is evident that this reduction technique favours the feature reduction, as its main criteria are based on the association rules discovered in the feature sets of the paths. We should also mention here, that in contrast to our preliminary work, we optimised the feature set extracted from the frequent patterns (Algorithm~\ref{alg:arred}, L.11-18) and now the reduction through association rules process is providing timely responses. This is also evident through our experiments in Section~\ref{sec:timeanalysis}.



\begin{algorithm}[ht]
  \SetAlgoLined
\SetKwInOut{Input}{input}\SetKwInOut{Output}{return}
\Input{$p_i$, $p_j$, $feature\_names$, $min\_max\_feature\_values$}
\Output{$similarity_{ij}$}
 $s_{ij} \gets 0 $\\
 \For{$f \in feature\_names$}{
  \uIf{$f \in p_i \wedge f \in p_j$}{
   find $l_i$, $u_i$, $l_j$, $u_j$ lower and upper bounds\\
   $inter \gets min(u_i,u_j) - max(l_i,l_j)$,
   $union \gets max(u_i,u_j) - min(l_i,l_j)$\\
   \lIf{$inter > 0 \wedge union \neq 0$}{$s_{ij} \gets s_{ij} + inter/union$}
   }\ElseIf{$f \notin p_i \wedge f \notin p_j$}{
   $s_{ij} \gets s_{ij} + 1 $\\
  }
 }
 \KwRet{$s_{ij}/|feature\_names|$}
 \caption{Path similarity metric}
 \label{alg:similarityMetric}
\end{algorithm}

\begin{algorithm}[ht]
 \KwIn{$paths, minimum\_number\_paths, clusters$}
 \KwOut{$reduced\_paths, reduced\_feature\_set$}
    $dissimilarity \gets []$ \\
    \For{$i \in [0,|paths|]$}{
        $vector \gets []$\\
        \For{$j \in [0,|paths|]$}{
            \eIf{$i = j$}{
                $vector \gets vector + [0]$
            }{
                $vector \gets vector + [compute\_dissimilarity(path[i], path[j])]$
            }
        }
        $dissimilarity \gets dissimilarity + [vector]$
    }
    
    $clusters = cl(dissimilarity, clusters)$\tcp*{$cr$ can be $k$-medoids, OPTICS, SC} 
    $sorted\_clusters = sort\_by\_size(clusters)$
    
    $reduced\_paths \gets []$, $k \gets 0$ \\
    \While{$|reduced\_paths| < minimum\_number\_paths \wedge k < |antecedent\_features|$}{
        \For{$path \in sorted\_cluster[k]$}{
            $reduced\_paths \gets reduced\_paths + path$
        }
        $k \gets k + 1$
    }
    $reduced\_feature\_set \gets \emptyset $\tcp*{Empty set of features}
    \For{$path \in reduced\_paths$}{
        \For{$f_i \in path$}{
            $reduced\_feature\_set \gets reduced\_feature\_set + f_i$
        }
    }
    \KwRet{$reduced\_paths,reduced\_feature\_set$}
 \caption{Reduction through clustering}
 \label{alg:clred}
\end{algorithm}

\subsubsection{Reduction through clustering}
A second reduction strategy based on clustering (CR) is utilised only on the binary and multi-class tasks. Reduction through clustering was not used in regression because reduction through association rules almost reaches the maximal local error allowed, and the overhead of clustering does not justify the effort\footnote{Reduction through clustering was not used in regression because reduction through association rules almost reaches the maximal local error allowed, and the overhead of clustering does not justify the effort}. Aside from $k$-medoids~\cite{kmedoidsori}, which was used in our preliminary work, OPTICS~\cite{OPTICS} and Spectral Clustering~\cite{SP} (SC) were added as additional choices. These algorithms are well-known clustering algorithms which need a distance or dissimilarity metric to find optimum clusters. Thus, the clustering of paths will require a distance or dissimilarity metric between two paths. Therefore, firstly, a similarity metric has been constructed (Algorithm~\ref{alg:similarityMetric}), in order to transform it to a dissimilarity metric, in a way that favours the absence of a feature from both paths. Specifically, if a feature is missing from both paths, the similarity of these paths increases by 1. When a feature is present in both paths, the similarity increases by a value between 0 and 1, i.e. the intersection of the two ranges normalised by the union of the two ranges. Finally, we subtract by 1 to convert the similarity metric, which has a range $[0,1]$, where 0 means the paths are not similar at all and 1 means the paths are identical, to a dissimilarity metric.

Using one of the aforementioned clustering algorithms, by default we use $k$-medoids, the clusters are estimated using the dissimilarity metric discussed above. Subsequently, the ordering of the clusters is performed on the basis of the number of paths they cover. Then, paths from larger clusters are accumulated into a list, until at least the required number of paths (as estimated in Section~\ref{mnpr}) has been obtained. 

By summing up larger clusters firstly, the probability of reducing the features is increasing, since the paths inside a cluster appear to be more similar among them. In addition, the biased dissimilarity metric will cluster paths with fewer insignificant features, leading to a subset of paths that are satisfied with a smaller set of features. As a result, we can assume that the reduction through clustering is also oriented toward feature reduction. The corresponding procedure is also represented in Algorithm~\ref{alg:clred}.


\subsubsection{Reduction through random selection}

Random selection (Algorithm~\ref{alg:rared}) (RS) of paths is used to achieve the minimum number of paths, in case of reduction through clustering, has not reached the quorum. For binary and multi-class tasks, RS removes randomly the unnecessary paths to only keep the minimum required number of paths. On the regression tasks, if the $local\_$ $error$ after the AR is smaller than the $allowed\_error$ RS is applied in order to minimise the paths, while maintaining the $local\_error$ less or equal to $allowed\_error$.

\begin{algorithm}[ht]
 \KwIn{$paths, minimum\_number\_paths/allowed\_error, task$}
 \KwOut{$reduced\_paths, reduced\_feature\_set$}
    $reduced\_paths \gets []$\\
    \eIf{$task \in \{binary, multi-class\}$}{
        \For{$i \in [0,minimum\_number\_paths]$}{
            $j \gets random(0,|paths|)$\\
            $reduced\_paths \gets reduced\_paths + paths[j]$\\
            $paths \gets paths - paths[j]$\\
        }
    }{
        $local\_error \gets 2*allowed\_error$\\
        $last\_path \gets \emptyset$\\
        $temp\_paths \gets paths$\\
        \While{$local\_error < allowed\_error \wedge |paths|>1$}{
            $j \gets random(0,|temp\_paths|)$\\
            $last\_path \gets temp\_paths[j]$\\
            $temp\_paths \gets temp\_paths - temp\_paths[j]$\\
            $local\_error \gets compute\_error(paths,temp\_paths)$
        }
        $reduced\_paths \gets temp\_paths + last\_path$
    }    

    $reduced\_feature\_set \gets \emptyset $\tcp*{Empty set of features}
    \For{$path \in reduced\_paths$}{
        \For{$f_i \in path$}{
            $reduced\_feature\_set \gets reduced\_feature\_set + f_i$
        }
    }
    \KwRet{$reduced\_paths,reduced\_feature\_set$}
 \caption{Reduction through random selection}
 \label{alg:rared}
 \end{algorithm}

\begin{algorithm}[ht]
 \KwIn{$instance, trees, allowed\_error, variation$}
 \KwOut{$reduced\_paths, reduced\_feature\_set, error$}
    $reduced\_paths \gets [tree..decision\_path(instance), \forall tree \in trees]$\\
    $error \gets 0$
    $predictions \gets [tree.predict(instance), \forall tree \in trees]$\\
    $mean \gets \frac{1}{|predictions|}\sum_{prediction \in predictions} prediction$\\
    $\sigma \gets  \sqrt{\frac{1}{|predictions|}\sum_{prediction \in predictions} (prediction-mean)^2}$\\
    \For{$s \in [.1,.2,.5,1,2,4,5,6,7,8,9,10,20,50,100]$}{
        $normal\_distribution \gets generate\_distribution(mean,\frac{\sigma}{s})$\\
        $normal\_min = min(normal\_distribution)$\\
        $normal\_max = max(normal\_distribution)$\\
        $local\_reduced\_paths \gets [], new\_prediction \gets 0$\\
        \For{$tree \in trees$}{
            $path \gets tree.decision\_path(instance)$, $prediction \gets tree.predict(instance)$\\
            \eIf{$(variation \in \{DSi\} \wedge \neg(prediction < normal\_min \vee prediction>normal\_max)) \vee (variation \in \{DSo\} \wedge (prediction <normal\_min or prediction>normal\_max))$}{
                $local\_reduced\_paths \gets local\_reduces\_paths + path$\\
                $new\_prediction \gets new\_prediction + prediction$
            }{
                $distance\_with\_min \gets |prediction - tree.min\_prediction|$\\
                $distance\_with\_max \gets |prediction - tree.max\_prediction|$\\
                \ElseIf{$distance\_with\_min > distance\_with\_max$}{
                    $new\_prediction \gets new\_prediction + tree.min\_prediction$
                }{
                    $new\_prediction \gets new\_prediction + tree.max\_prediction$
                }
            }
            $local\_error \gets |mean - \frac{1}{|trees|}new\_prediction|$\\
            \If{$local\_error < allowed\_error \wedge |reduced\_paths|>|local\_reduces\_paths| $}{}
                $error \gets local\_error, reduced\_paths \gets local\_reduced\_paths$
        }
    }

    $reduced\_feature\_set \gets \emptyset $\tcp*{Empty set of features}
    \For{$path \in reduced\_paths$}{
        \For{$f_i \in path$}{
            $reduced\_feature\_set \gets reduced\_feature\_set + f_i$
        }
    }
    \KwRet{$reduced\_paths,reduced\_feature\_set,error$}
 \caption{Reduction through distribution-based selection}
 \label{alg:dsred}
 \end{algorithm}

\subsubsection{Reduction through distribution-based selection}\label{sec:reddbs}
The last technique, reduction through distribution-based selection (DS), is a reduction technique exclusively designed for regression tasks. This method utilises normal distributions, is not combined with any other method, and it attempts to reduce the features and the paths of a rule for a given instance. Each instance's prediction is the average of the individual predictions of the trees. For example, in an RF regression model with 100 trees, an instance's prediction will be the average of the 100 predictions by each one of trees. Those 100 predictions have their own distribution. 

\begin{figure}[ht]
\centerline{\includegraphics[width=0.9\textwidth]{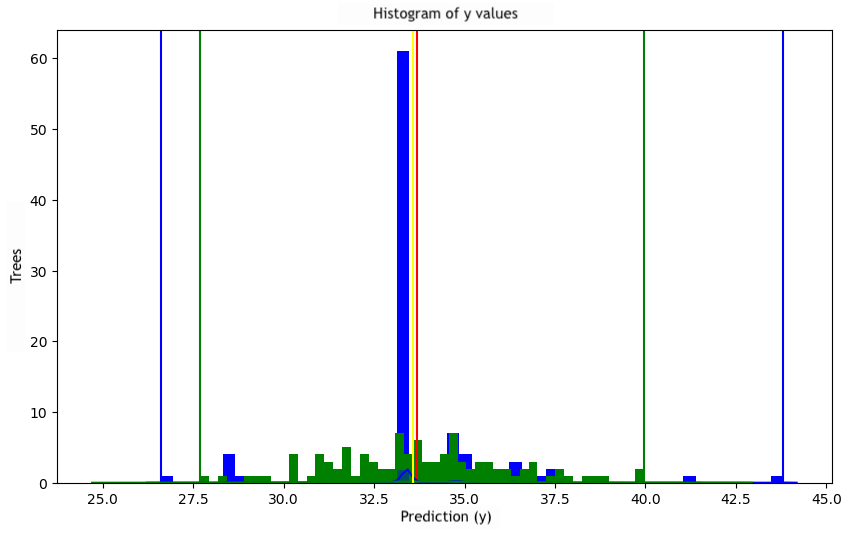}}
\caption{Distribution of predictions of a given instance
} \label{regrDistr}
\end{figure}

In Figure~\ref{regrDistr} the actual distribution of the RF's prediction is shown in blue, while the green distribution is a normal distribution generated with the mean and standard deviation ($\sigma$) values of the original distribution. We can change the way this artificial distribution is created by multiplying or dividing the $\sigma$ value, generating wider or narrower distributions around the mean value. Therefore, for different variations of the $\sigma$ value, we apply one of the following techniques, inner or outer selection, to choose the paths we will keep for the final rule. We select the distribution's $\sigma$ value, which achieves the lowest number of paths with a $local\_error \in [0, allowed\_error]$ for an instance interpretation. This process is also presented in Algorithm~\ref{alg:dsred}, which is based on Algorithm~\ref{alg:localerror}.

\paragraph{Inner selection} The distribution-based inner selection (DSi) technique, for all the different $\sigma$ values, it selects the paths inside the range of the generated distribution to form the interpretation rule. For the paths outside the generated distribution, we compute the $local\_error$. This approach is going to reduce the number of paths, when the majority of trees are providing similar predictions close to the mean value. Then, by isolating and removing distant predictions, even if they shift, the mean will not be greatly affected, and thus the $local\_error$ will be relatively small.

\paragraph{Outer selection} Similarly to the inner selection, the distribution-based outer selection (DSo) technique, for all the different $\sigma$ values, it selects the paths outside the range of the generated distribution to form the interpretation rule. Again, for the paths outside the generated distribution, the $local\_error$ is computed. This approach is providing better results when the original distribution is either like an inverse normal distribution or uniform. In this case the majority of the predictions will be distant to the mean value. Keeping those predictions, and leaving out from the interpretation rule the predictions closer to the mean, we will probably have a low $local\_error$, because the predictions closer to the mean will be fewer. Moreover, those predictions are less likely to deviate a lot when small changes to the input will occur, thus the error most of the time will be smaller than the estimated $local\_error$.




\subsection{Feature-ranges formulation}
\label{frext}
In this section, we present how the formulation of feature-ranges are taking place in a binary classification scenario. However, in all tasks the formulation procedure is the same. 

Consider an RF of $T$ trees that predicts instance $x$ as $c_j \in C =\{0,1\}$. We concentrate on the $K \geq \frac{|T|}{2} + 1$ trees that classify $x$ as $c_j$. For each feature we calculate the range of values imposed by the conditions in the root-to-leaf path of each of the $K$ trees in which it appears.

\begin{figure}[!htb]
\centering
\minipage{0.49\textwidth}
\centering
\includegraphics[width=1\textwidth]{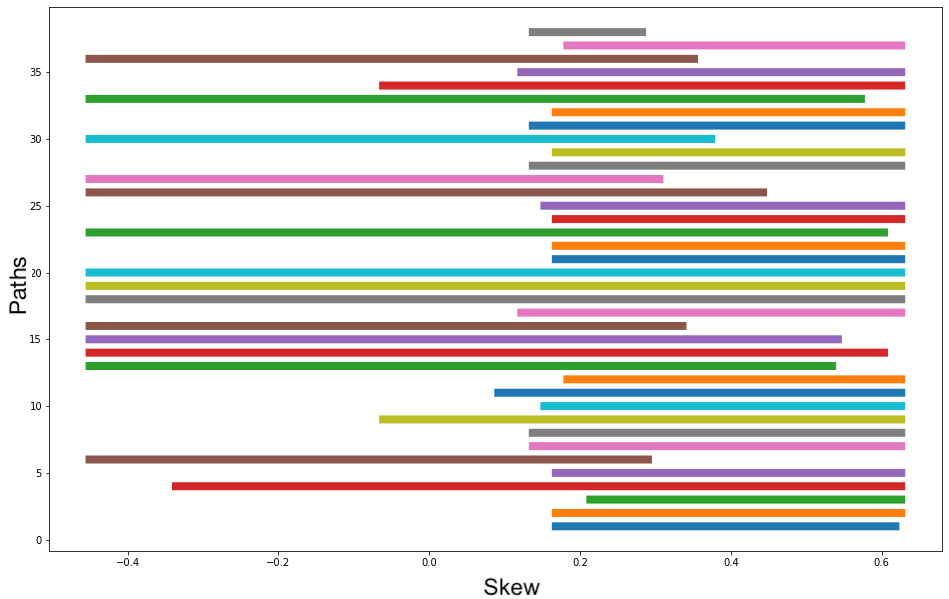}
\caption{Bars plot of `skew' feature} \label{BarsPlot}
\endminipage \hspace{0.05in}
\minipage{0.49\textwidth}
\centering
\includegraphics[width=1\textwidth]{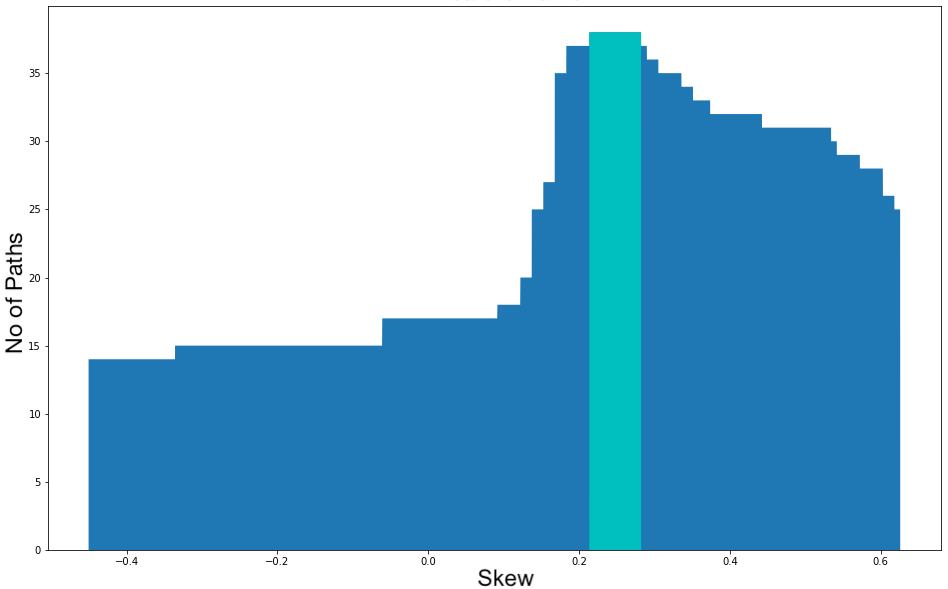}
\caption{Stacked area plot of `skew' feature} \label{stackedAreaPlot}
\endminipage\hfill
\end{figure}

For a concrete real-world example, we train an RF with $T=50$ trees on the Banknote dataset~\cite{ucidata}. We focus on the {\em skew} feature of Banknote and a test instance $x$, whose {\em skew} value is 0.25. The decision of the RF for this instance is supported by $K=39$ of its trees. Figure~\ref{BarsPlot} presents the value ranges for {\em skew} in each of the 38 trees, whose decision paths contain the {\em skew} feature. Figure~\ref{stackedAreaPlot} presents the corresponding stacked area plot, which shows the number of paths in the $y$ axis whose value ranges include the corresponding value in the $x$ axis.

The cyan/light grey region of Figure~\ref{stackedAreaPlot} is the intersection between these 38 decision paths for {\em skew}. Then, we could say that the range of this feature will be: $0.21 \leq skew = 0.25 \leq 0.029$. This intersection will always contain the value ($0.25$) for the particular feature of the examined instance. Furthermore, no matter how much the value of the feature may shift, the decision paths will not change as long as it remains within this intersection. Thus, if the instance's value for {\em skew} changes from $0.25$ to $0.28$ each tree will take its decision via the same path. Summing up the above the interpretation may have the following shape:

\begin{quote}
\centering
    \smaller{`if $0.21 \leq skew \leq 0.29$ and $\dots$ then {\em Fake Banknote}'}
\end{quote}

These paths will be used in order to form a final single rule. However, using all the available trees will produce interpretations with many features, thus less comprehensible to the end-user. Moreover, a large number of paths can lead to a small, strict, and very specific feature ranges. For example, the {\em skew} value of the instance $x$ was $0.25$ and the intersection range of all paths for this feature is $0.21\leq skew \leq 0.29$, while the global range of {\em skew} is $[-1,1]$. If the 10 first paths corresponding to the first 10 top rows (top to bottom) of the stacked area plot (Figure~\ref{stackedAreaPlot}) could have been reduced, the range would get $0.15$$\leq f_1$$\leq 0.4$, thus a broader one. A narrow range, like the one mentioned above, would result in a negative perception of the model, which would be considered inconsistent and unreliable, whereas a wider range would be less rebuttable and preferable. Hence this is the reason why we compute the minimum number of paths required~\ref{mnpr}, and then we apply reduction techniques~\ref{redTechs}.


In this work, we also developed a much faster way for computing these intersections. 
This affected the responsiveness of any operation, including the extraction of an explanation without reduction. The following assumptions about the performance are confirmed by the time analysis studies in Section~\ref{sec:timeanalysis}.

\subsection{Categorical features}
\label{subsec:catf}
One more LF's adaptation concerns the handling of categorical features. Focusing only in OneHot encoded cases, we designed a handling procedure, which is in accordance with this principle: if the OneHot encoded feature appears in the reduced paths, then the categorical feature where it originates is added to the rule. If this does not exist in the reduced paths, this means that the prediction has not been affected by this value of the categorical feature. However, we are looking for the other OneHot encoded features of the same categorical feature appearing in the reduced rules, but not in the instance (their values are equal to zero), and we present them to the user, along with the interpretation, as categorical values that may change the prediction if the category changes to one of them.

To give a brief example, let us have a feature $height$ with values $ = [short, normal,$ $tall]$. After OneHot encoding, there are three features: $height_{short}$, $height_{normal}$ and $height_{tall}$. An instance with $height = short$ will therefore have $height_{short}$ $=1$, $height_{normal}$ $=0$, and $height_{tall}$ $=0$ after encoding. If the feature $height_{short}$ appears in the extracted paths, it means that it influences the prediction and will appear in the final rule as `$height=short$'. Elsewhere, this indicates that this feature value does not affect the prediction. Then, we check whether the features $height_{normal}$ and $height_{tall}$ appear in the reduced paths. If any of them appears on the paths, this means that they influence the prediction if they change. Therefore, they are provided to the user as alternative values.

\subsection{Interpretation composition}
The last stage of LionForests combines the extracted ranges of features and the information derived from the categorical handling, if there was any categorical feature, in a single natural language rule. The order of appearance of the ranges of features in the rule is determined by a method of global interpretations, called permutation importance~\cite{randomForests}. An example of an interpretation is as follows:

\begin{equation}
\centering
    \text{`if } 0 \leq f_1 \leq 0.5 \text{ and } -0.5 \leq f_2 \leq 0.15 \text{ and } f_3 = \text{{\em short}} \text{ then {\em Prediction}'.}
\end{equation}

This rule can be interpreted in this way: ``As long as the value of the $f_1$ is between the ranges 0 and 0.5, and the value of $f_2$ is between the ranges -0.5 and 0.15, and $f_3$ equals {\em sort}, the system will predict this instance as {\em Prediction}. If the value of $f_1$, $f_2$ or both, surpass the limits of their ranges, and $f_3$ changes category then the prediction may change. Note that the features are ranked through their influence''. This type of interpretation is intuitive, human-readable and more comprehensible than other interpretations, like feature importance. Furthermore, this rule is conclusive since it will always include all of the feature ranges appearing in the paths produced the specific prediction, while these paths will be greater than or equal to the minimum number of paths required to maintain the prediction stable.

\subsection{Visualisation}
Apart from extending LionForests to multi-class and regression tasks, as well as optimising the core functionalities of LionForests, we are also presenting a proposed visualisation layout. Using this interface, the user will see the explanation rule for a prediction, and then the ability to choose to inspect each of the features that appear in the rule will be available. In the case of a feature with numerical values, the user will be supplied with a distribution plot with the values of the feature. Two vertical green lines are being used to represent the range that appears in the rule, while two blue vertical lines reflect the ranges of the original rule, without applying any reduction technique. On the other hand, if the selected feature is categorical, then if alternative values are available as described in Section~\ref{subsec:catf}, they will be visible to the user. An example is provided in Section~\ref{sec:visexample} and Figures~\ref{ex:ex1} and~\ref{ex:ex2}.

\section{Evaluation}
\label{sec:exp}
The performed experiments were related to response time, sensitivity analysis to inspect how RF parameters affect feature and path reduction, a comparative study with similar well-known techniques, as well as a qualitative assessment with examples. LionForests' code and evaluation experiments are available at GitHub repository ``LionLearn''\footnote{\url{https://git.io/JY0gF}}.

\subsection{Setup and datasets}
For the series of experiments, we have used the Scikit-learn's~\cite{sklearn} RandomForestClassifier and RandomForestRegressor. We used 8 different datasets, 3 for binary, 2 for multi-class, 2 for regression and 1 for multi-class and regression problems. The datasets for binary tasks are: Banknote~\cite{ucidata}, Heart (Statlog)~\cite{ucidata} and Adult (Census)~\cite{adultDataset}. Glass~\cite{ucidata} and Image Segmentation (Statlog)~\cite{ucidata} for multi-class tasks. For regression, we used Wine Quality~\cite{wineDataset} and Boston Housing~\cite{bostonDataset}. Finally, Abalone~\cite{abaloneDataset} was used for both multi-class and regression. 

We applied a 10-fold cross-validation grid search for each dataset using the following set of parameters: \textit{max depth} $\in$ \{1, 5, 7, 10\} (max depth of each individual tree), \textit{max features} $\in$ \{`sqrt', `log2', 75\%, None = all features\} (max number of features per individual tree), \textit{min samples leaf} $\in$ \{1, 2, 5, 10\}, \textit{bootstrap} $\in$ \{True, False\} (bootstrap samples or the whole dataset is used to train each tree), \textit{estimators} $\in$ \{10, 100, 500, 1000\} (number of trees in the ensemble). The scoring metric of the grid search was the weighted $F_1$-score for the binary and multi-class datasets, and the mean absolute error ($mae$) for the regression datasets. Table~\ref{tab:parametersAndScores} shows the number of instances and features of each dataset, the parameters' values that achieved the best $F_1$-score/$mae$ for each dataset, along with the scores itself.

\begin{table}[ht]
\centering
\resizebox{\textwidth}{!}{%
\begin{tabular}{ccccccc|c|c|c|c|c|}
\cline{8-12} & & & & & &  & \multicolumn{5}{c|}{Parameters}\\ \hline
\multicolumn{1}{|c|}{Task} & \multicolumn{1}{c|}{Dataset} & \multicolumn{1}{c|}{Classes} & \multicolumn{1}{c|}{Instances} & \multicolumn{1}{c|}{Features} & \multicolumn{1}{c|}{Categorical} & \begin{tabular}[c]{@{}c@{}}$F_1$-score/\\ $mae$\end{tabular} & \begin{tabular}[c]{@{}c@{}}No. \\ Estimators\end{tabular} & \begin{tabular}[c]{@{}c@{}}Max \\ Depth\end{tabular} & \begin{tabular}[c]{@{}c@{}}Max \\ Features\end{tabular} & \begin{tabular}[c]{@{}c@{}}Min \\ Samples\\ Leaf\end{tabular} & Bootstrap \\ \hline
\multicolumn{1}{|c|}{\multirow{3}{*}{\begin{tabular}[c]{@{}c@{}}Binary\\ (B)\end{tabular}}} & \multicolumn{1}{c|}{Banknote} & \multicolumn{1}{c|}{2} & \multicolumn{1}{c|}{1372} & \multicolumn{1}{c|}{4} & \multicolumn{1}{c|}{-} & 99.49\% & 500 & 10 & 0.75 & 1 & True\\ \cline{2-12} 
\multicolumn{1}{|c|}{} & \multicolumn{1}{c|}{Heart (statlog)} & \multicolumn{1}{c|}{2} & \multicolumn{1}{c|}{270} & \multicolumn{1}{c|}{13} & \multicolumn{1}{c|}{1} & 84.60\% & 500 & 5 & `sqrt' & 5 & False\\ \cline{2-12} 
\multicolumn{1}{|c|}{} & \multicolumn{1}{c|}{Adult Census} & \multicolumn{1}{c|}{2} & \multicolumn{1}{c|}{45167} & \multicolumn{1}{c|}{12 (85)} & \multicolumn{1}{c|}{7 (80)} & 84.84\% & 1000 & 10 & 0.75 & 2 & True\\ \hline
\multicolumn{1}{|c|}{\multirow{2}{*}{\begin{tabular}[c]{@{}c@{}}Multi-Class\\ (MC)\end{tabular}}} & \multicolumn{1}{c|}{Glass} & \multicolumn{1}{c|}{6} & \multicolumn{1}{c|}{214} & \multicolumn{1}{c|}{9} & \multicolumn{1}{c|}{-} & 71.95\% & 1000 & 10 & `sqrt' & 2 & True\\ \cline{2-12} 
\multicolumn{1}{|c|}{} & \multicolumn{1}{c|}{I. Segmentation} & \multicolumn{1}{c|}{7} & \multicolumn{1}{c|}{2310} & \multicolumn{1}{c|}{19} & \multicolumn{1}{c|}{-} & 96.98\% & 500 & 10 & 0.75 & 1 & False\\ \hline
\multicolumn{1}{|c|}{MC/R} & \multicolumn{1}{c|}{Abalone} & \multicolumn{1}{c|}{4/-} & \multicolumn{1}{c|}{4027} & \multicolumn{1}{c|}{8} & \multicolumn{1}{c|}{-} & 68.76\%/1.44 & 10/100 & 10 & 0.75 & 10/5 & True\\ \hline
\multicolumn{1}{|c|}{\multirow{2}{*}{\begin{tabular}[c]{@{}c@{}}Regression\\ (R)\end{tabular}}}  & \multicolumn{1}{c|}{Boston}  & \multicolumn{1}{c|}{-} & \multicolumn{1}{c|}{506} & \multicolumn{1}{c|}{13} & \multicolumn{1}{c|}{-} & 2.95 & 1000 & 10 & 0.75 & 1 & True\\ \cline{2-12} 
\multicolumn{1}{|c|}{} & \multicolumn{1}{c|}{Wine Quality} & \multicolumn{1}{c|}{-} & \multicolumn{1}{c|}{4898} & \multicolumn{1}{c|}{12} & \multicolumn{1}{c|}{-} & 0.55 & 1000 & 10 & 0.75 & 5 & True\\ \hline
\end{tabular}%
}
\caption{Performance and best parameters for each dataset}
\label{tab:parametersAndScores}
\end{table}

\subsection{Sensitivity analysis}
\label{sec:sensanalysis} 
We have conducted the following sensitivity analysis experiments in order to compare the capability of LF to reduce features (feature reduction ratio - FR\%) and paths (path reduction ratio - PR\%) while tuning both the parameters of LF (LF) and the configuration of the Random Forests (RF) model. In order to simplify the information obtained from the analysis, we present each task (binary, multi-class, regression) separately. The parameters of the RF model under examination are the \textit{max depth} $\in$ \{1, 5, 7, 10\}, \textit{max features} $\in$ \{`sqrt', `log2', 75\%, None\} and \textit{estimators} $\in$ \{10, 100, 500, 1000\}. LF parameters tested in this sensitivity analysis are the \textit{AR algorithm} (AR) $\in$ \{apriori, fpgrowth\}, \textit{CL algorithm} (CR) $\in$ \{kmedoids, OPTICS, CS\}, and \textit{method} $\in$ \{1, 2, 3, 12, 13, 23, 123\} for the binary and multi-class classification, where 1 represents the use of AR, 2 the use of CL and 3 the use of RS. Therefore, 123 means the utilisation of all 3 methods, in this order. For regression, we used \textit{method} $\in$ \{AR+RS, DSi, DSo\}, while an additional parameter of LF, $allowed\_error$ is also examined in the regression tasks. 

\subsubsection{Binary classification}

\begin{figure}[ht]
\centerline{\includegraphics[width=1\textwidth]{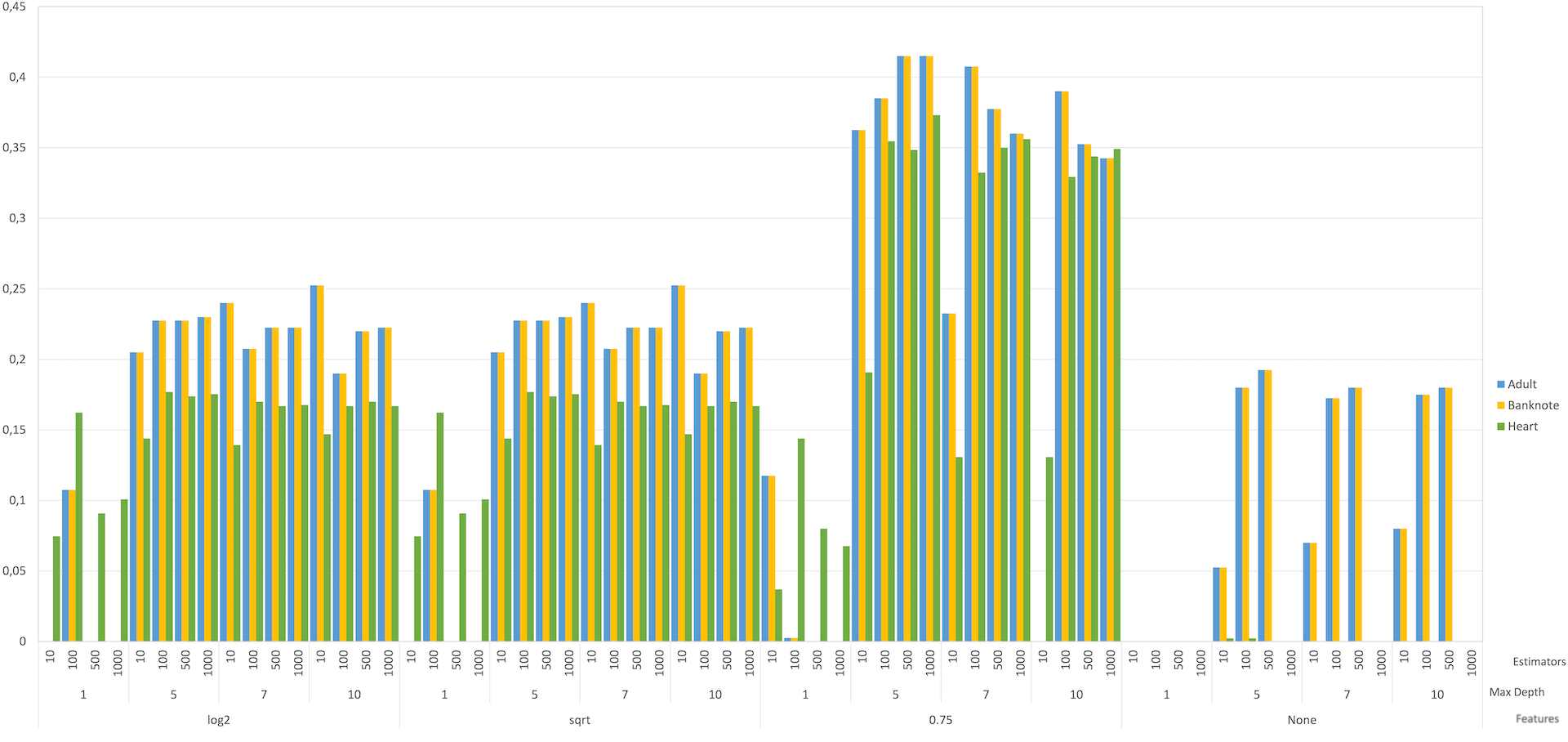}}
\caption{Binary classification: analysis of FR relation to RF's parameters} \label{bc1}
\end{figure}

\begin{figure}[ht]
\centerline{\includegraphics[width=1\textwidth]{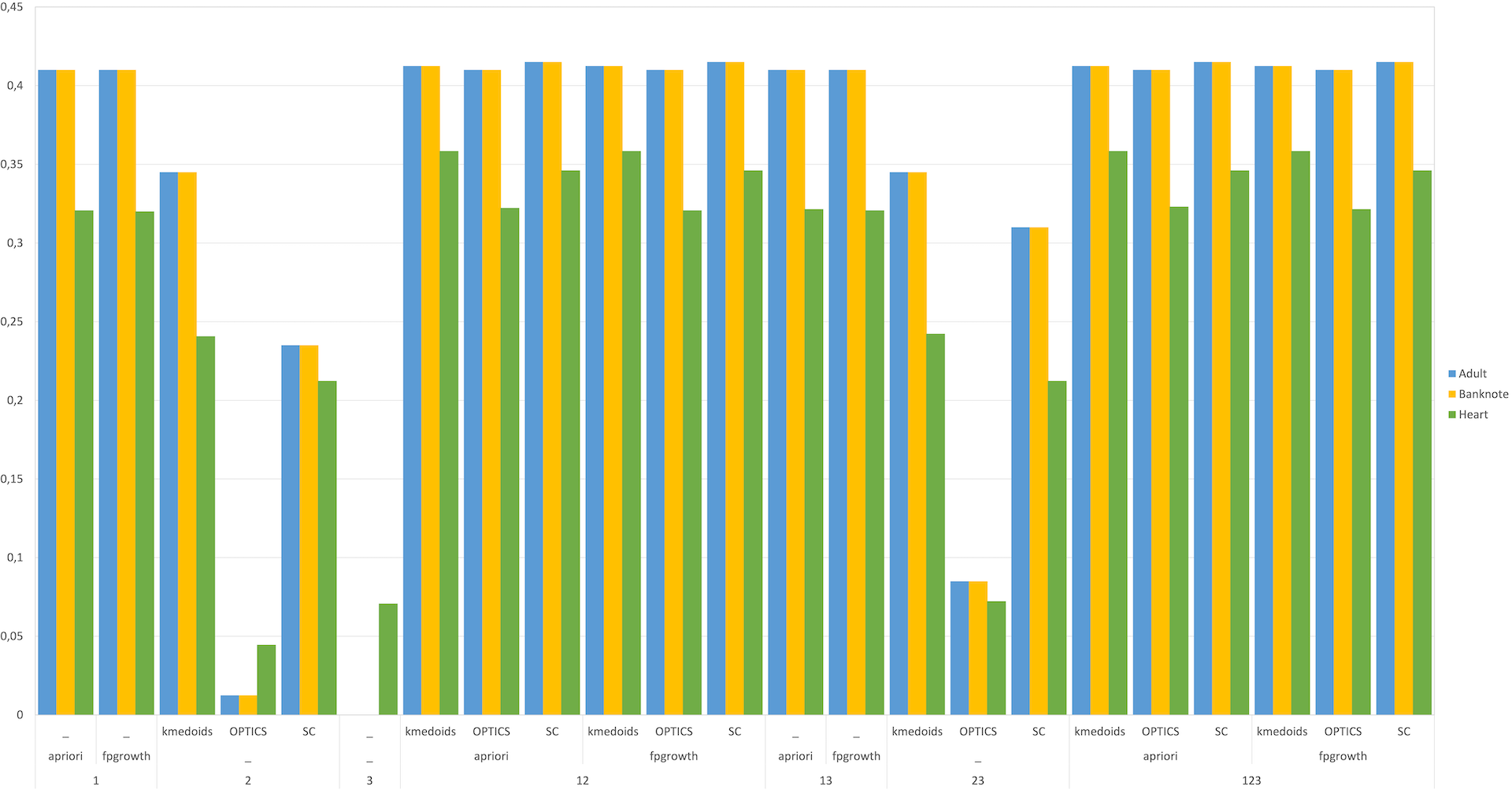}}
\caption{Binary classification: analysis of FR relation to LF's parameters} \label{bc2}
\end{figure}

The first sensitivity analysis we carried out applies to the Adult, Banknote and Heart (Statlog) datasets. Figure~\ref{bc1} presents the FR\% for the parameters of the RF while Figure~\ref{bc2} refers to the parameters of LF. The parameter analysis reveals that when the RF's \textit{max features} parameter is set to 75\%, the reduction in features in all datasets is higher. With regard to \textit{max depth} and \textit{estimators}, LF achieves over 35\% FR when the \textit{max depth} is greater than or equal to $5$ and the \textit{estimators} are $100$ or more.

In Figure~\ref{bc2}, we can see how the parameters of LF affect the FR\% in these datasets. We initially observe that the two different AR (1) are performing identically in the FR\%. In CR (2), the FR\% seemed to diverge for the different algorithms. Specifically, $k$-medoids and SC manage to reduce the features by 20\% or more in all datasets, while OPTICS could not manage to perform any reduction. RS (3) did not achieve any FR in Adult and Banknote, while it achieved low FR\% in Heart (Statlog). Among the three approaches, AR seems necessary to achieve high FR\%, while the combination of AR (1) with CL (2) seems to increase in all three datasets the FR\%.

\begin{figure}[ht]
\centerline{\includegraphics[width=1\textwidth]{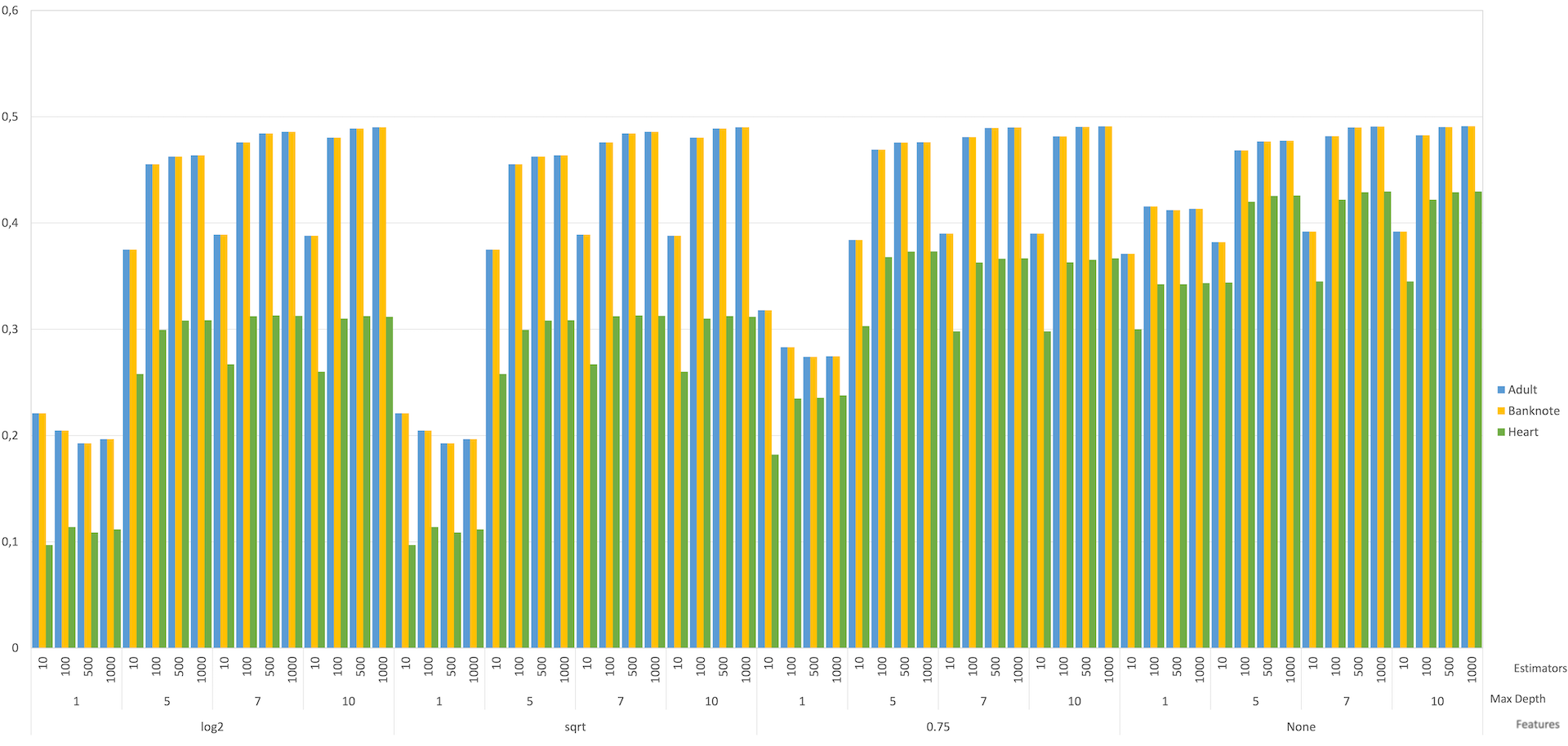}}
\caption{Binary classification: analysis of PR relation to RF's parameters} \label{bc3}
\end{figure}

About the PR analysis, Figure~\ref{bc3} reveals that when the RF's \textit{max features} parameter is set to `None', the reduction in paths in all datasets is higher. With regards to \textit{max depth} and \textit{estimators}, LF achieves over 40\% PR when the \textit{max depth} is greater or equal to $5$ and the \textit{estimators} are $100$ or over.

\begin{figure}[ht]
\centerline{\includegraphics[width=1\textwidth]{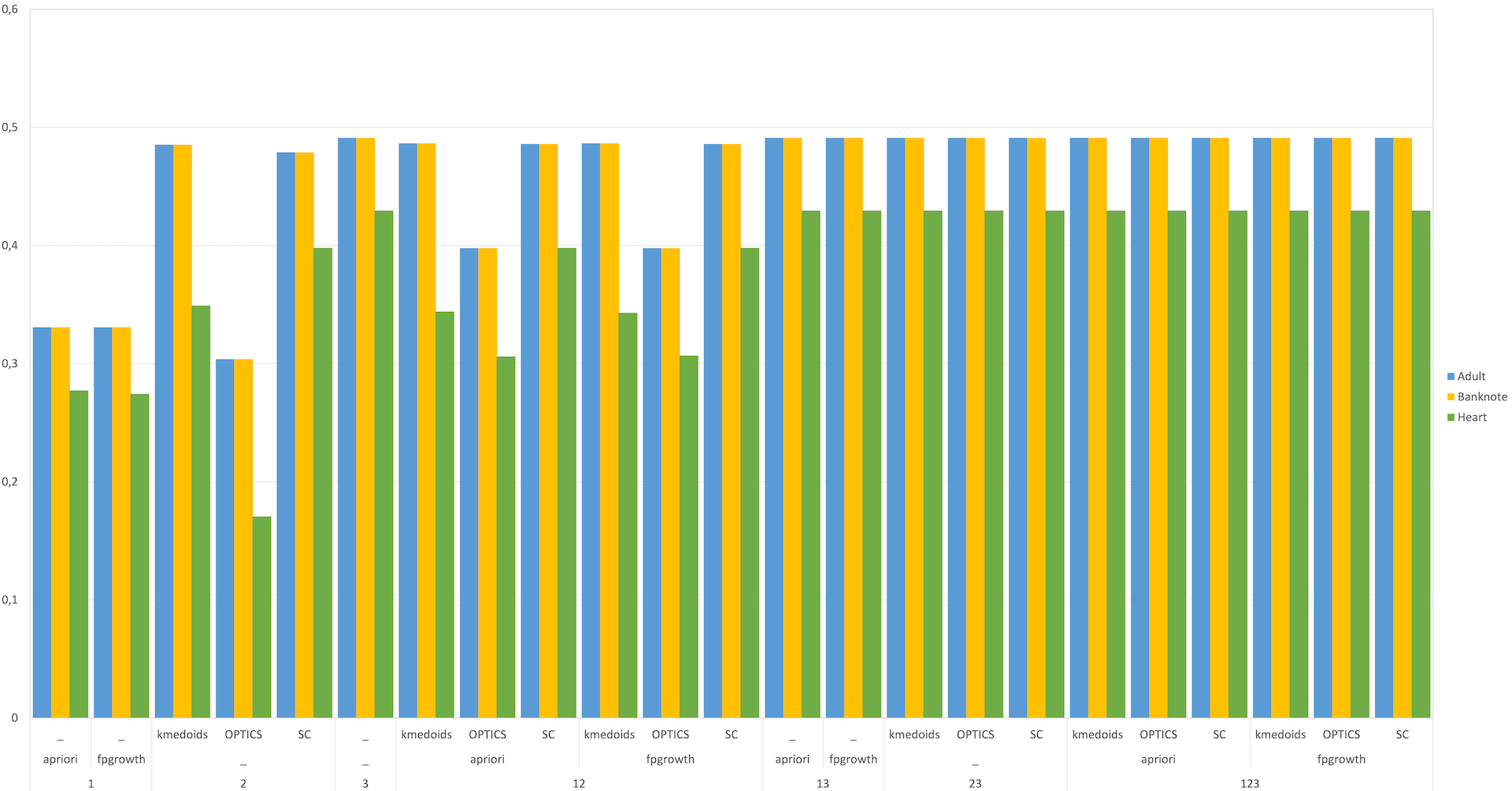}}
\caption{Binary classification: analysis of PR relation to LF's parameters} \label{bc4}
\end{figure}

In Figure~\ref{bc4}, we can see how the parameters of LF affect the PR\% in these datasets. In contrast to FR, for PR, CR (2) and RS (3) are both maximising the PR\%. Recall that we cannot reduce more than a quorum in a binary setup, thus these techniques achieving 49\% PR are performing optimally. AR (1), on the other hand, cannot seem to be able to optimally reduce paths. Finally, we observe that when combining all three techniques (123), for every parameter setting, the PR\% is higher than 40\%.

\subsubsection{Multi-class classification}

\begin{figure}[ht]
\centerline{\includegraphics[width=1\textwidth]{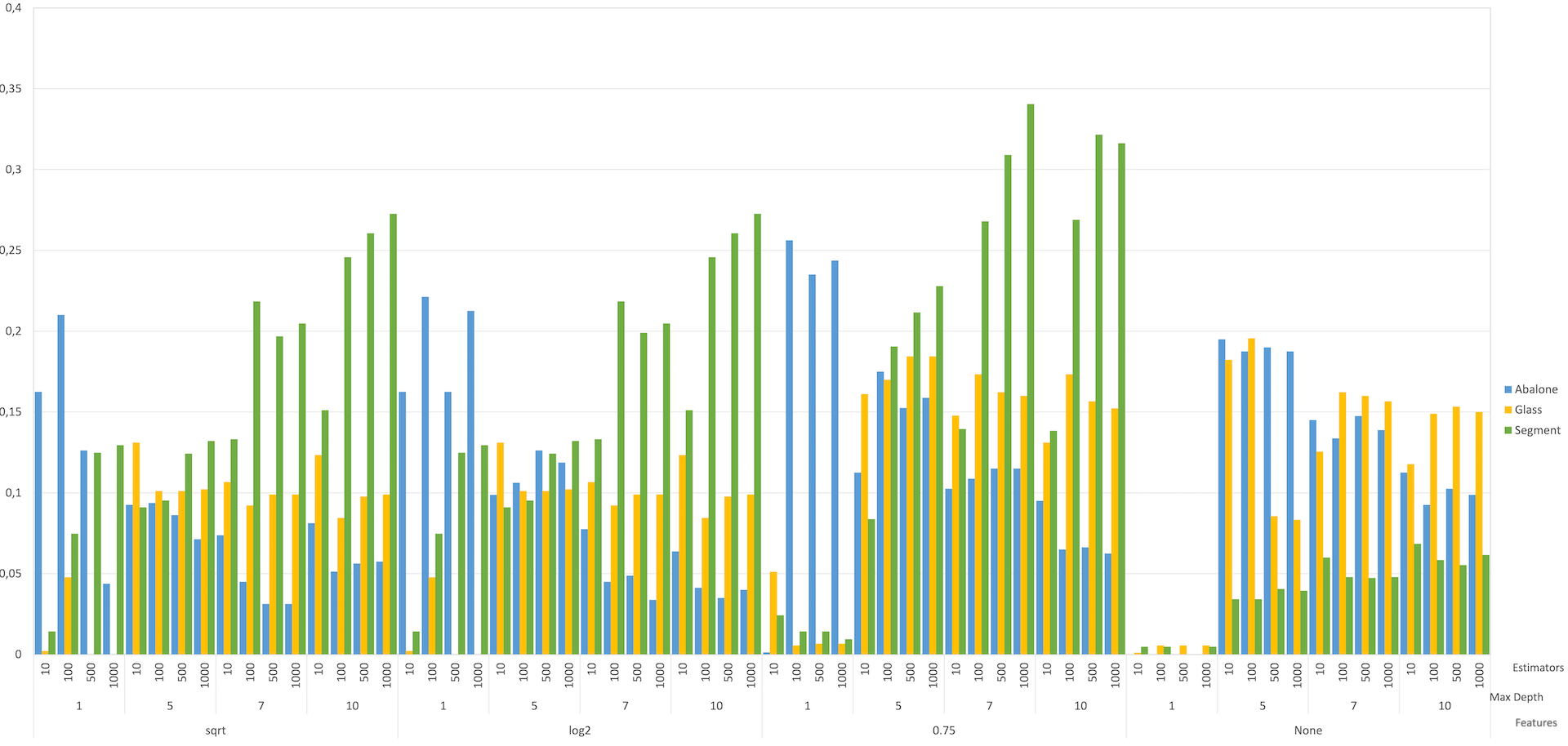}}
\caption{Multi-class classification: analysis FR relation to RF's parameters} \label{mc1}
\end{figure}

We continue with the multi-class datasets Glass, I. Segmentation and Abalone. The tuning of RF's parameters and their impact to the FR\% are visible in Figure~\ref{mc1}. The analysis reveals that when the RF's \textit{max features} parameter is set to 75\%, the FR in all datasets is higher. LF achieves over 17\% FR when the \textit{max depth} is greater than or equal to $5$ and \textit{estimators} are 100 or over, while it achieves more than 25\% and 34\% for the individual datasets, Abalone and I. Segmentation, respectively.

\begin{figure}[ht]
\centerline{\includegraphics[width=1\textwidth]{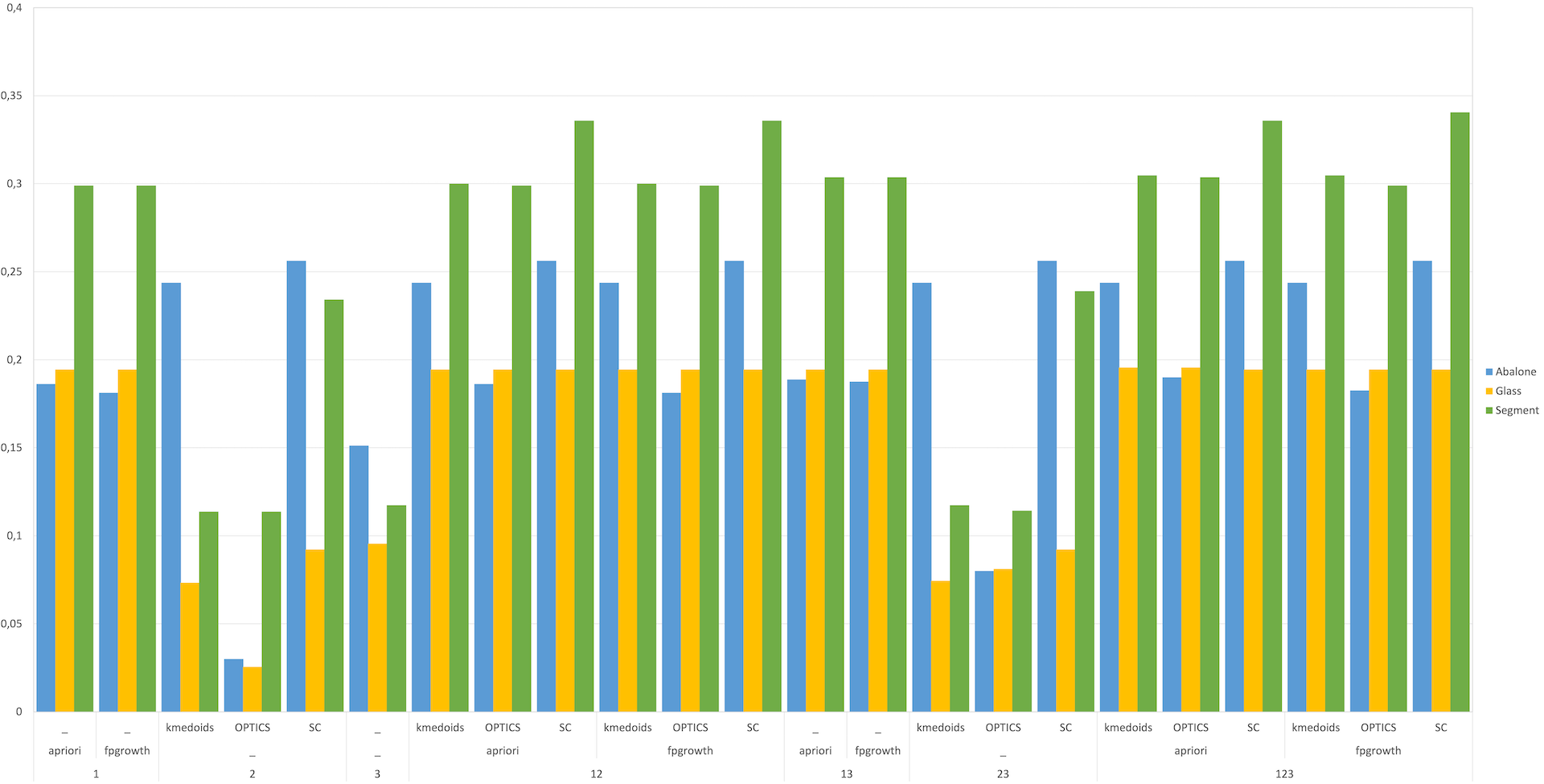}}
\caption{Multi-class classification: analysis of FR relation to LF's parameters} \label{mc2}
\end{figure}

Figure~\ref{mc2} presents how FR\% is affected based on the different parameters of LF. As observed in the binary classification sensitivity analysis, here as well it is eminent that AR (1) is performing equally in the FR\%. CR (2) in the Abalone dataset achieved a higher FR than AR, and when combined (123) with the other techniques, AR and RS, the FR\% is not increasing. The analysis of Glass dataset revealed that rather than the FR achieved by the AR (1), no other method or combination managed to increase the FR\%. Finally, on I. Segmentation seemed the combination of AR and CR (12), with specifically SC, to provide the highest FR\%. Another interesting point is that the RS (3) managed to reduce the features of the rules in all three datasets in contrast to the RS's performance on the binary's classification sensitivity analysis.

\begin{figure}[ht]
\centerline{\includegraphics[width=1\textwidth]{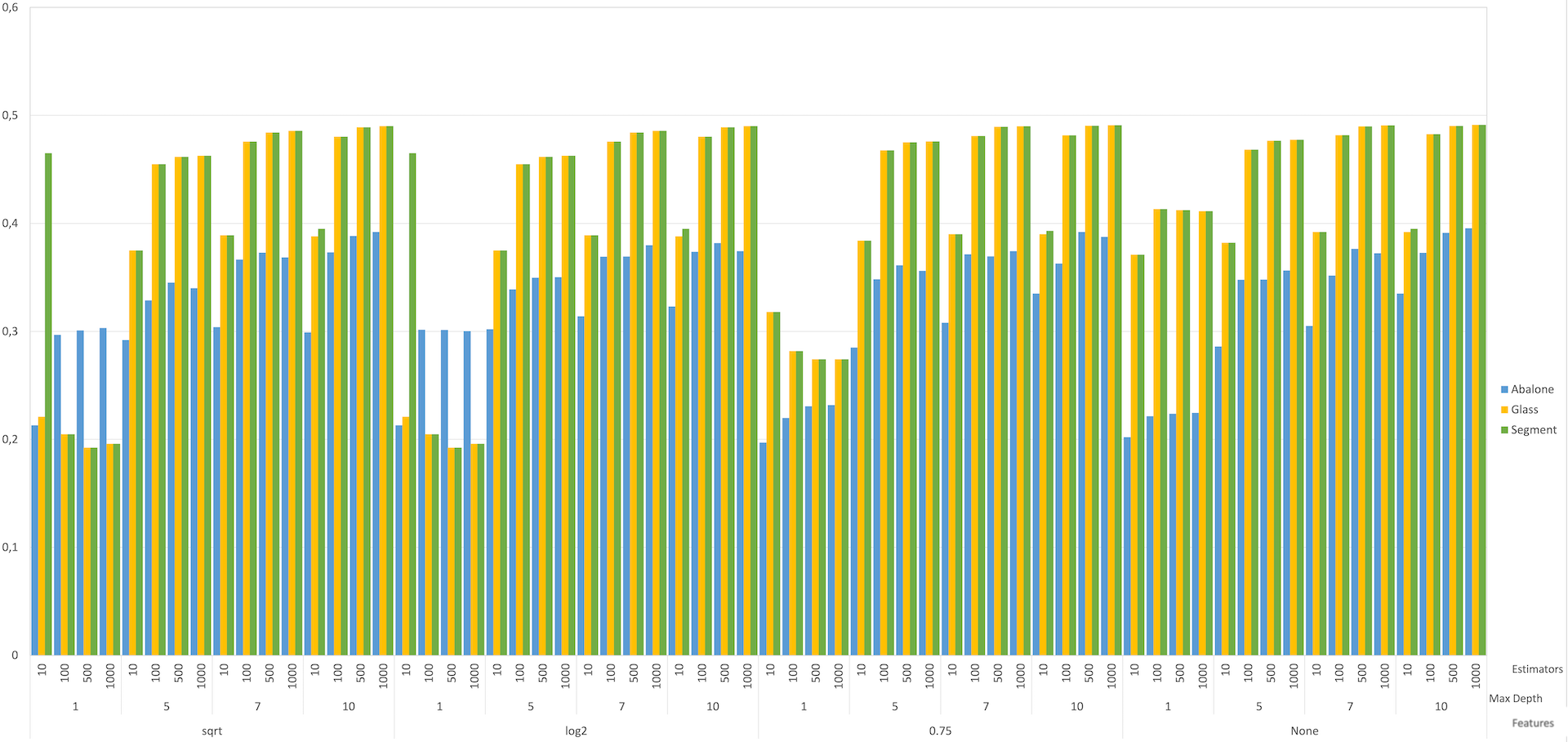}}
\caption{Multi-class classification: analysis of PR relation to RF's parameters} \label{mc3}
\end{figure}

\begin{figure}[ht]
\centerline{\includegraphics[width=1\textwidth]{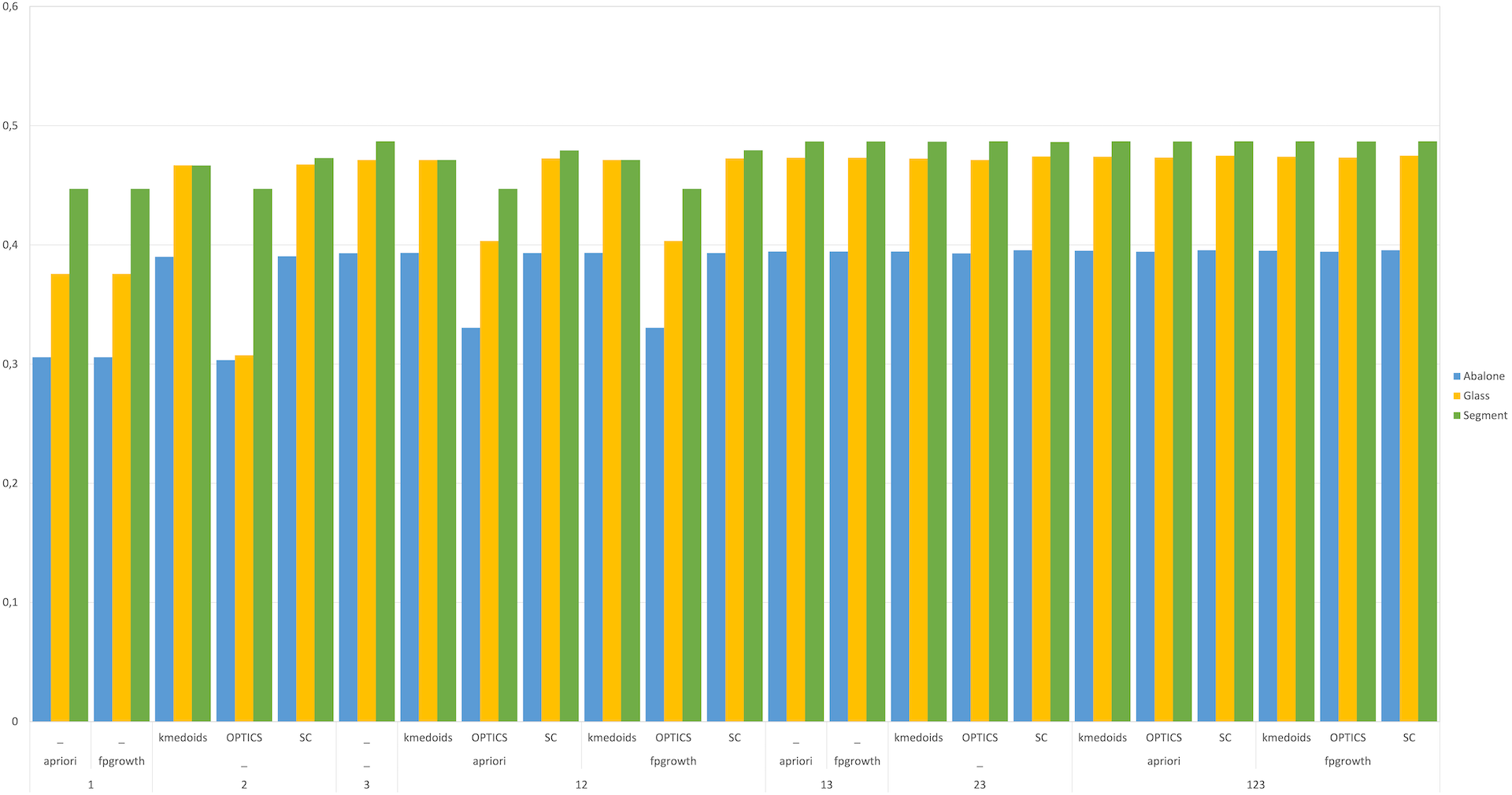}}
\caption{Multi-class classification: analysis of PR relation to LF's parameters} \label{mc4}
\end{figure}

Through Figure~\ref{mc3}, observing the PR while tuning the RF's parameters in these datasets, we can say that the \textit{max features} parameters do not affect the PR\%. We can not conclude the same for \textit{max depth} and \textit{estimators}, where we need 5 or higher and 100 or more, respectively, to achieve higher PR\%. The highest PR\% it is achieved when \textit{max depth} equals 10 and \textit{estimators} equals 1000.

In Figure~\ref{mc4}, we can see how the parameters of LF affect the PR\% in these datasets. RS (3) is maxing out the PR\%. AR (1) cannot seem to achieve the desirable PR results, while CL (2) is performing well, but not as good as RS (3). Thus, RS or any combination with RS leads to a PR\% of 38\% or more.

\subsubsection{Regression}
\begin{figure}[ht]
\centerline{\includegraphics[width=1\textwidth]{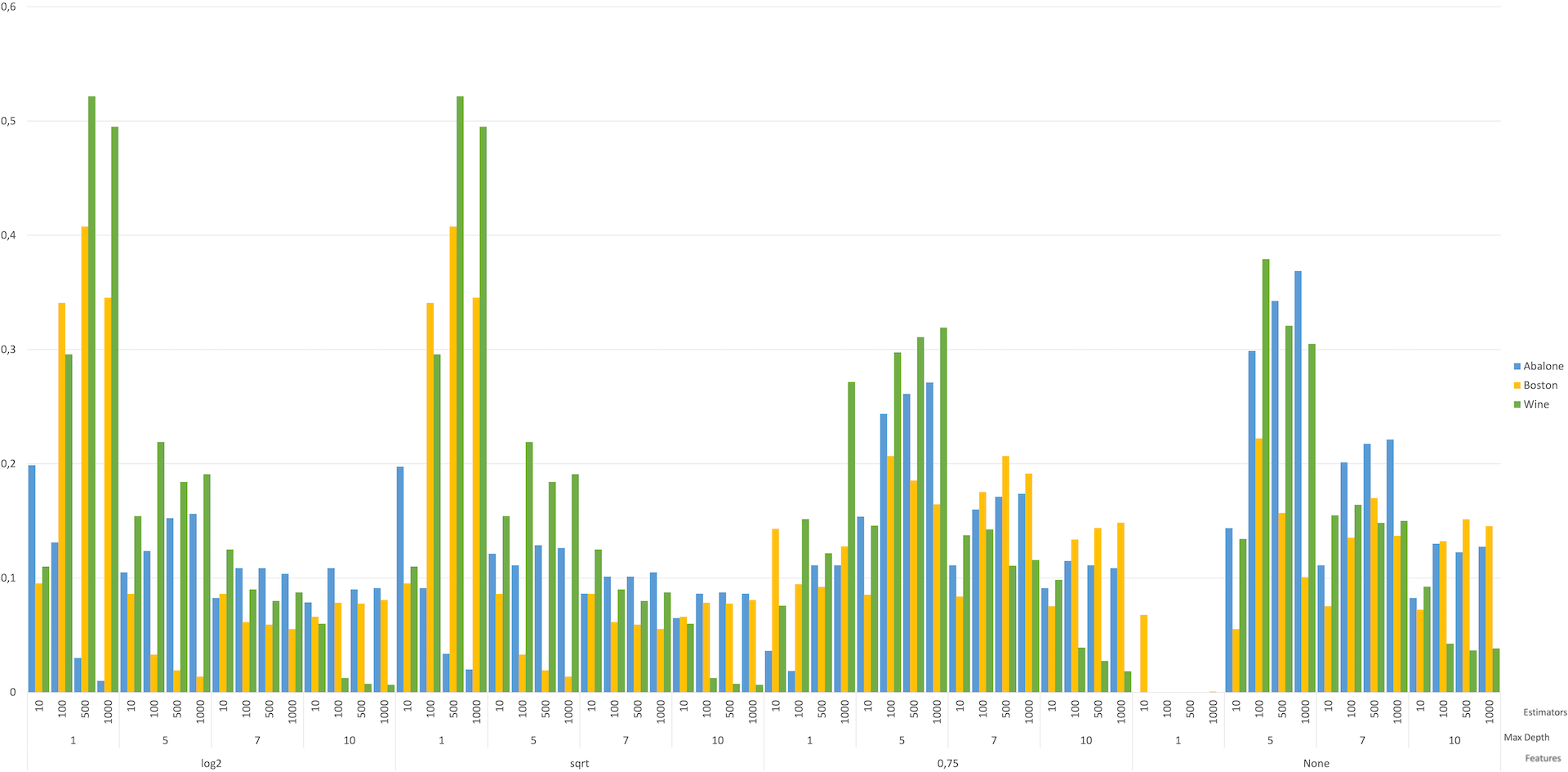}}
\caption{Regression: analysis FR relation to RF's parameters} \label{r1}
\end{figure}

The last set of experiments is related to the analysis of the regression datasets, Abalone, Boston and Wine. In Figure~\ref{r1}, the relation of RF's parameters to the FR\% is visible. We can say that the most influencing parameter is \textit{estimators}. When \textit{estimators} are equal or more than $500$ and \textit{max depth} is either 1 or 5, then the reduction is between $35\%$ to $51\%$. Moreover, for Boston and Wine we observe that when \textit{max features} is set to either `sqrt' or `log2', the FR\% is higher. On the other hand, higher \textit{max features} values like `0.75' or `None' seem to favour the FR\% for Abalone.

\begin{figure}[!htb]
    \centering
    \begin{minipage}{.4\textwidth}
        \centering
        \centerline{\includegraphics[width=1\textwidth]{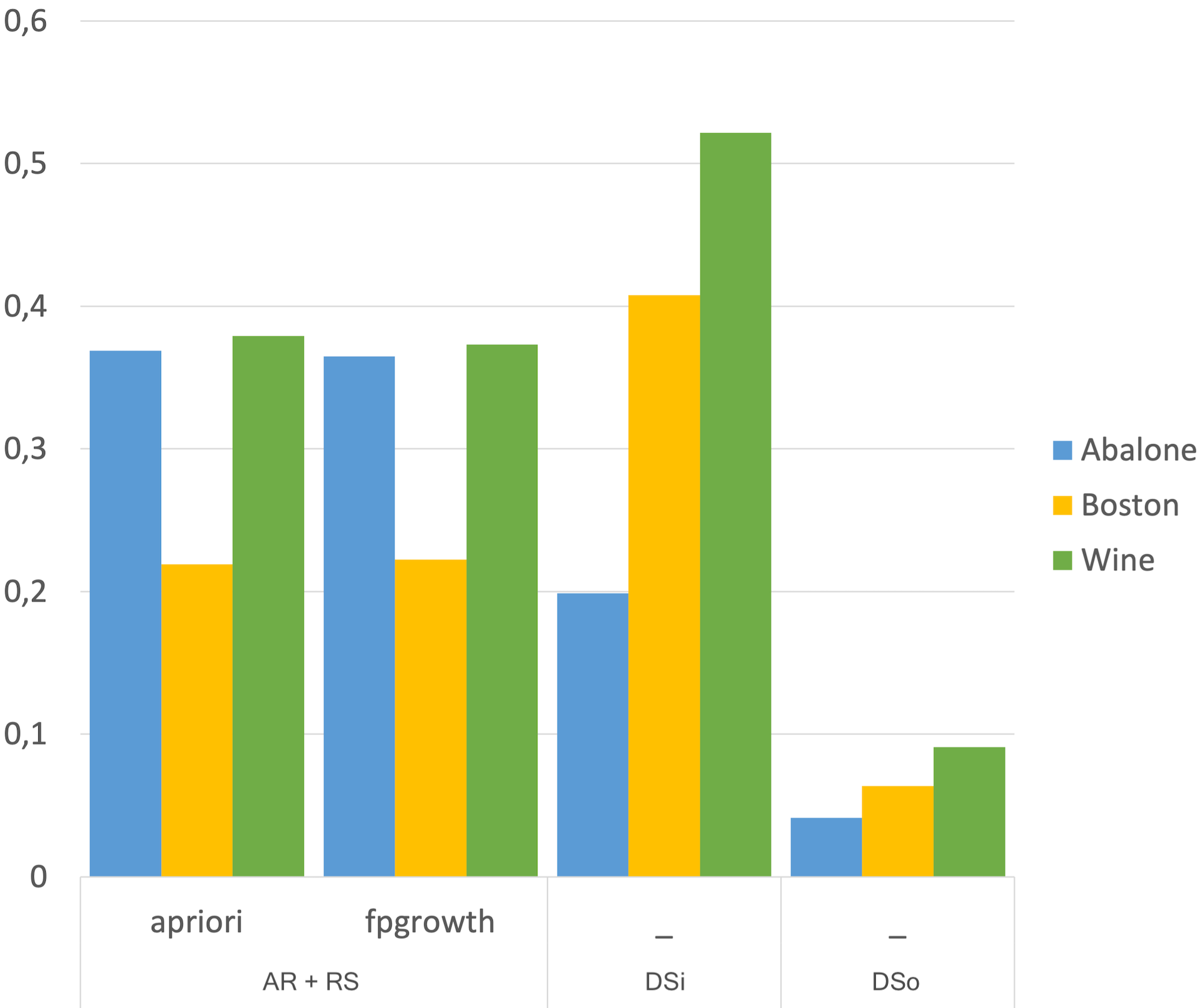}}
        \caption{Regression: analysis of FR relation to LF's parameters} \label{r2}
        \label{fig:prob1_6_2}
    \end{minipage}%
    \hspace{0.04\textwidth}
    \begin{minipage}{0.405\textwidth}
        \centering
        \centerline{\includegraphics[width=1\textwidth]{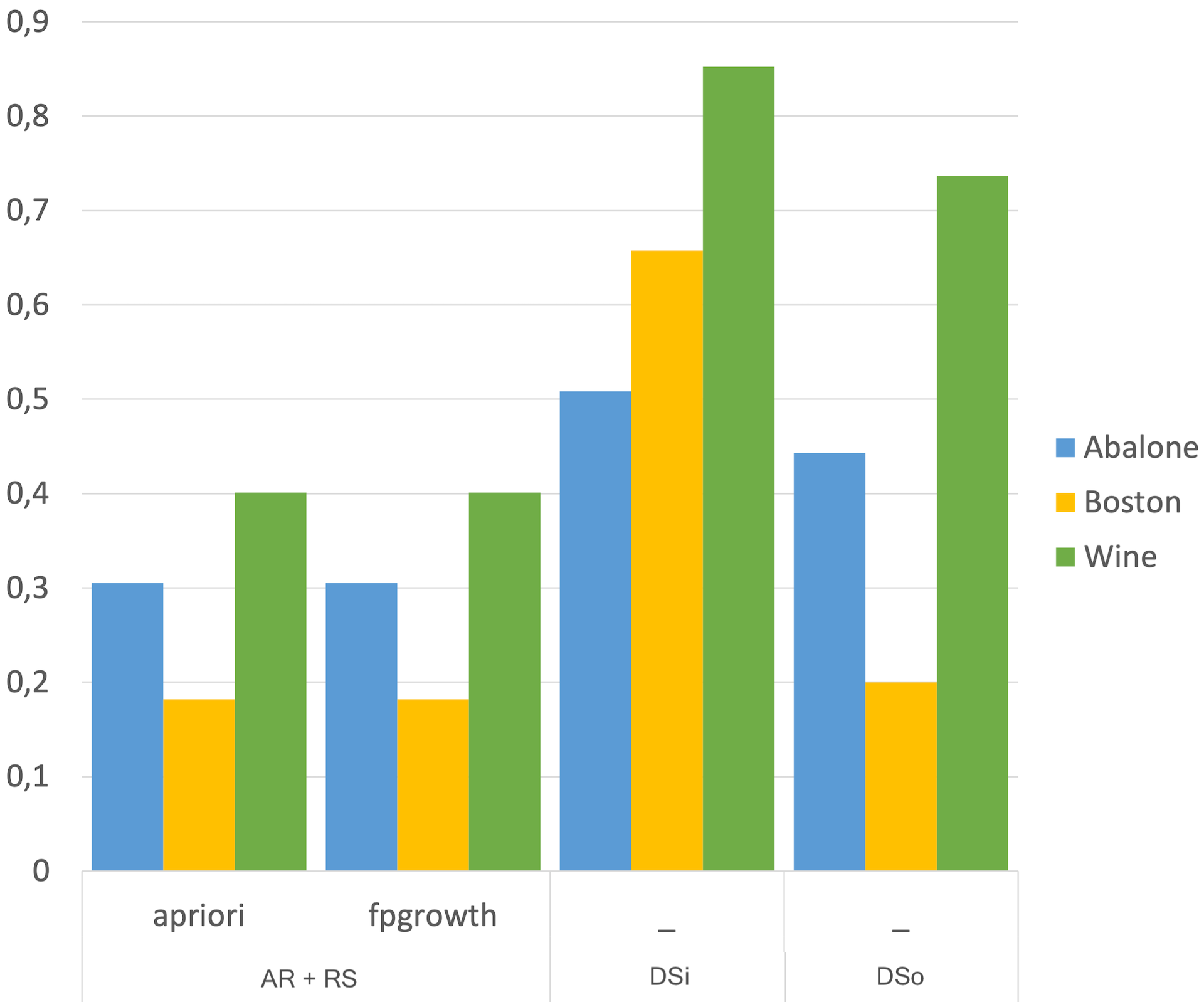}}
        \caption{Regression: analysis of PR relation to LF's parameters} \label{r4}
    \end{minipage}
\end{figure}

Inspecting how the LF's parameters are affecting the FR\%, in Figure~\ref{r2}, we can see that AR+RS method provides better results for Abalone, while DSi for Boston and Wine. However, DSo cannot reach desirable levels of FR\% for any case. 

\begin{figure}[ht]
\centerline{\includegraphics[width=1\textwidth]{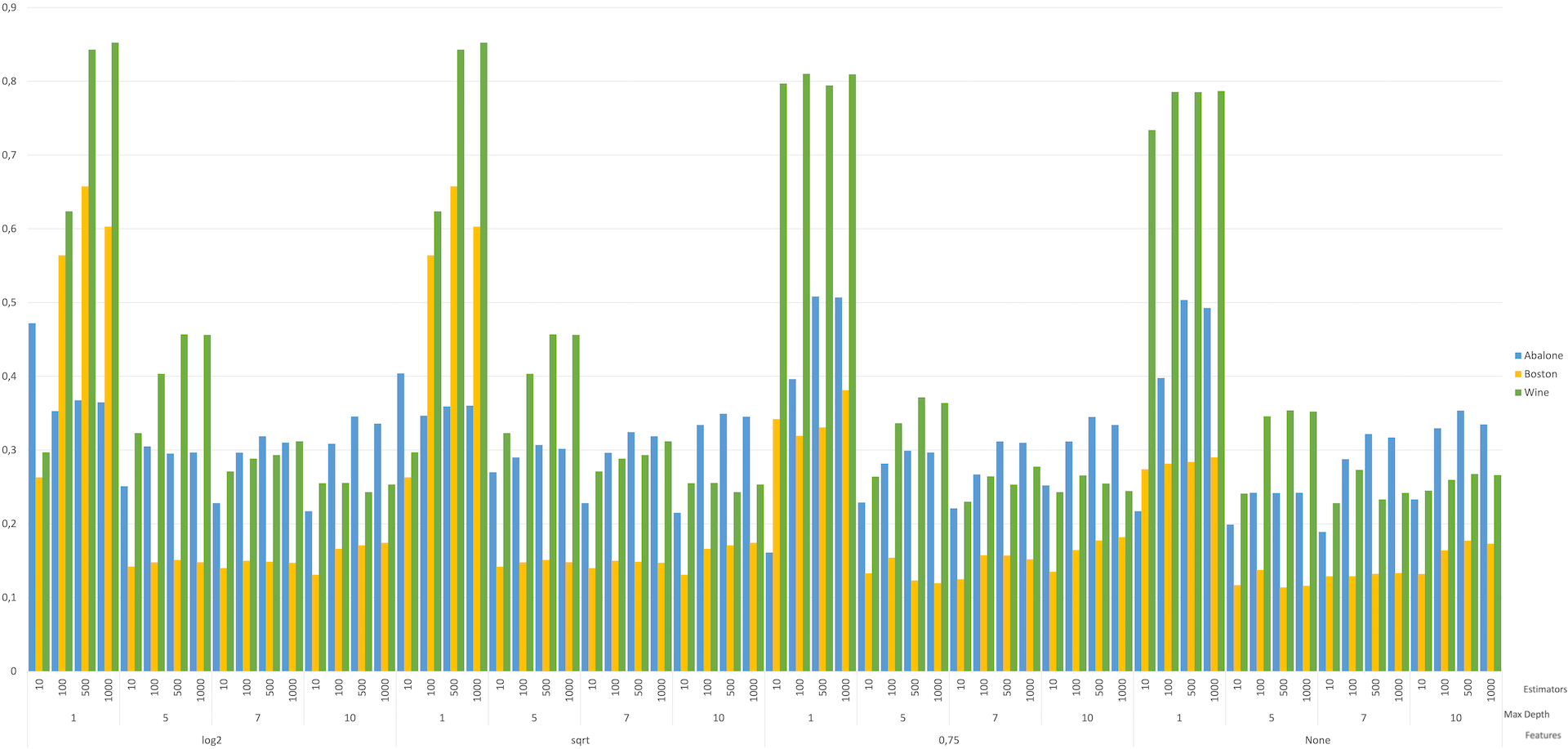}}
\caption{Regression: analysis of PR relation to RF's parameters} \label{r3}
\end{figure}

The same pattern we identified for the FR\% relation to RF's parameters, it is apparent for the relation of PR\% with the RF's parameters as well (Figure~\ref{r3}). Setting \textit{estimators} between $500$ or $1000$ and \textit{max depth} to either 1 or 5, the PR is between $50\%$ to $85\%$. However, \textit{max features} do not affect the PR\%.

In Figure~\ref{r4}, we can see how the parameters of LF affect the PR\% in these datasets. We observe the highest PR\%, over 50\%, with the DSi reduction method of LF. DSo is also better than AR+RS in terms of PR\%.

\begin{figure}[ht]
\centerline{\includegraphics[width=1\textwidth]{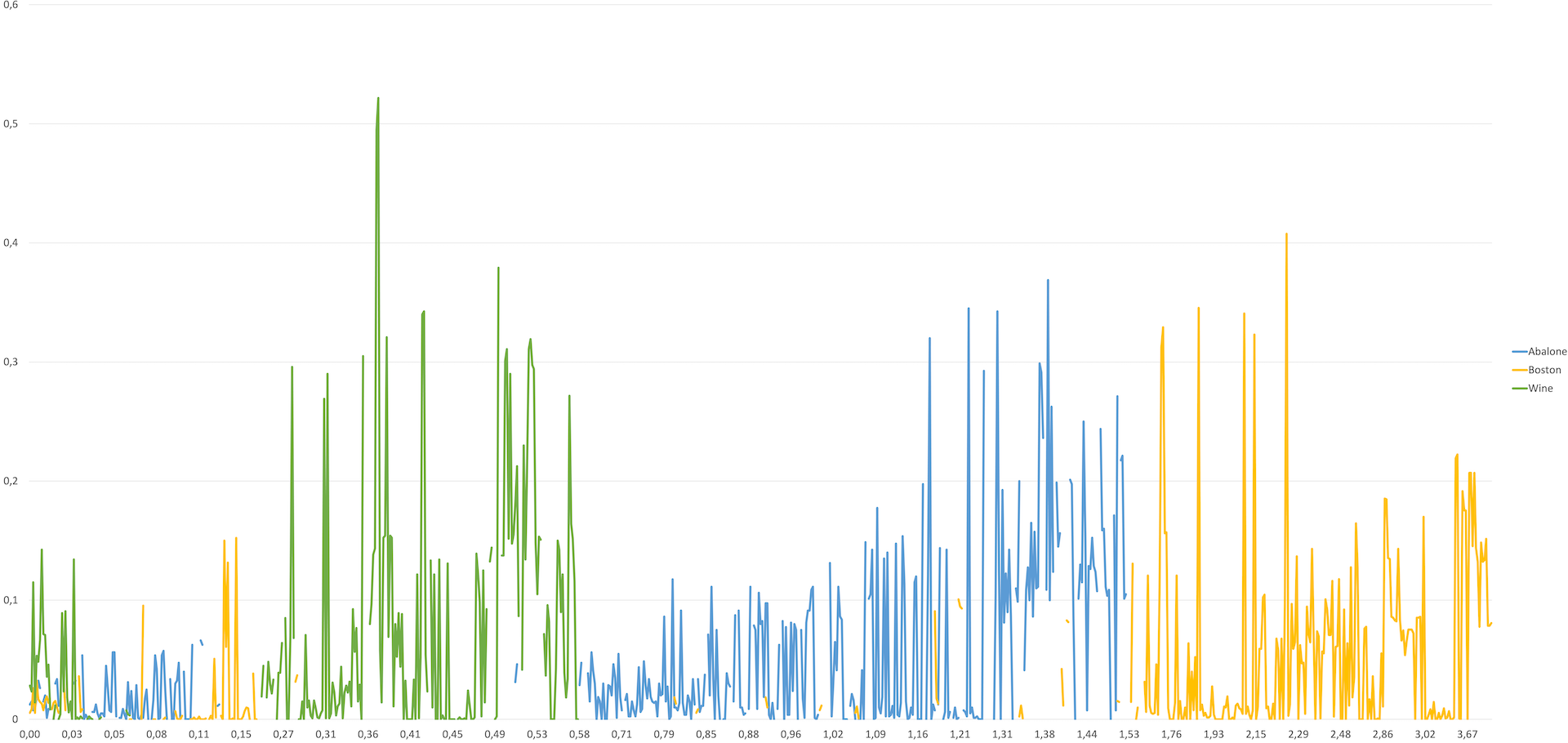}}
\caption{Regression: analysis of FR relation to $local\_error$} \label{r5}
\end{figure}

Examining the relation of $local\_error$ with the FR\% (Figure~\ref{r5}), we can say that for the Wine dataset we can achieve high FR\%, over 50\%, with a low $local\_error$ of around $0.36$. For the Abalone dataset, we need a $local\_error$ with a value between $[1.1, 1.4]$ in order to achieve approximately 35\% of FR. Finally, for Boston in order to achieve FR\% higher than 40\% we need a $local\_error$ around $2.2$. 

\begin{figure}[ht]
\centerline{\includegraphics[width=1\textwidth]{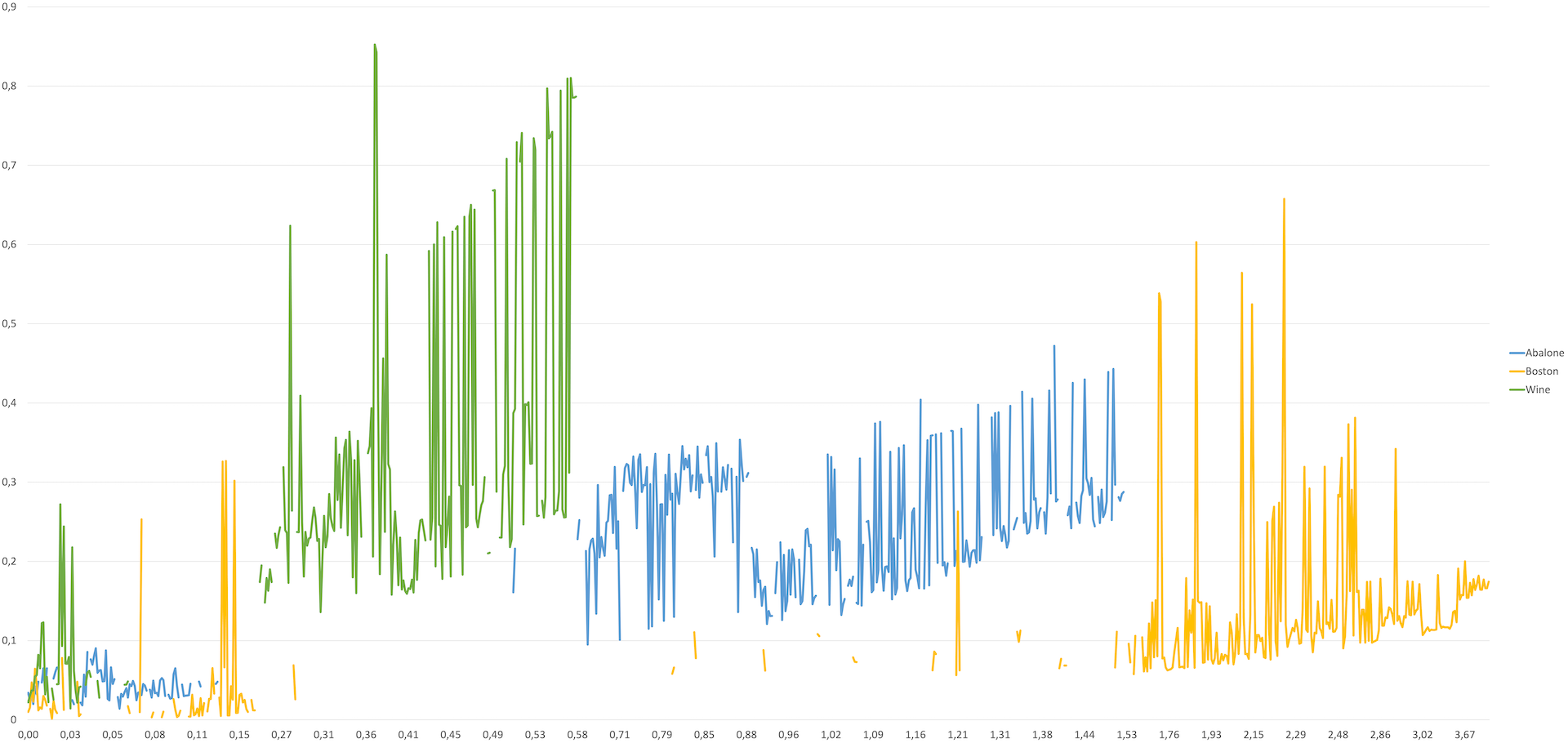}}
\caption{Regression: analysis of PR relation to $local\_error$} \label{r6}
\end{figure}

\begin{table}[ht]
\centering
\begin{tabular}{l|l|l|l|l|}
\cline{2-5}
                              & Min & Max  & Mean  & Std  \\ \hline
\multicolumn{1}{|l|}{Abalone} & 3.0 & 19.0 & 9.74  & 2.86 \\ \hline
\multicolumn{1}{|l|}{Boston}  & 5.0 & 50.0 & 22.53 & 9.19 \\ \hline
\multicolumn{1}{|l|}{Wine}    & 3.0 & 9.0  & 5.82  & 0.87 \\ \hline
\end{tabular}
\caption{Statistics of target variable of regression task's datasets}
\label{regr:stats}
\end{table}

Finally, about the relation of PR\% with the $local\_error$, we observe, in Figure~\ref{r6}, that we acquire higher PR\% when we allow higher $local\_error$, in every dataset. In order to let the reader understand better the relation of both the FR\% and PR\% with the $local\_error$, we present the target variable statistics of each dataset in Table~\ref{regr:stats}. This will help associating the $local\_error$ with the actual values of the target variable of each dataset.

\subsection{Time analysis}
\label{sec:timeanalysis}

Optimisation of LF implementation was an emerging issue to address, mainly due to the need for timely responses. The speed of providing explanations to a model's predictions in a variety of applications is indeed a very important factor, specifically, for on-line applications
. The second direction of our experiments targets the analysis of response time. We are comparing LF to its preliminary version in order to back up the assumption of Sections~\ref{subsec:rar} and~\ref{frext}. For this analysis we use only the binary datasets of Banknote and Heart (statlog) because the preliminary version of LF is only applicable to binary tasks, and we do not use the Adult dataset, as the old version was unable to finish our exhaustive analysis within a relatively reasonable time.

\begin{figure}[ht]
\centerline{\includegraphics[width=1\textwidth]{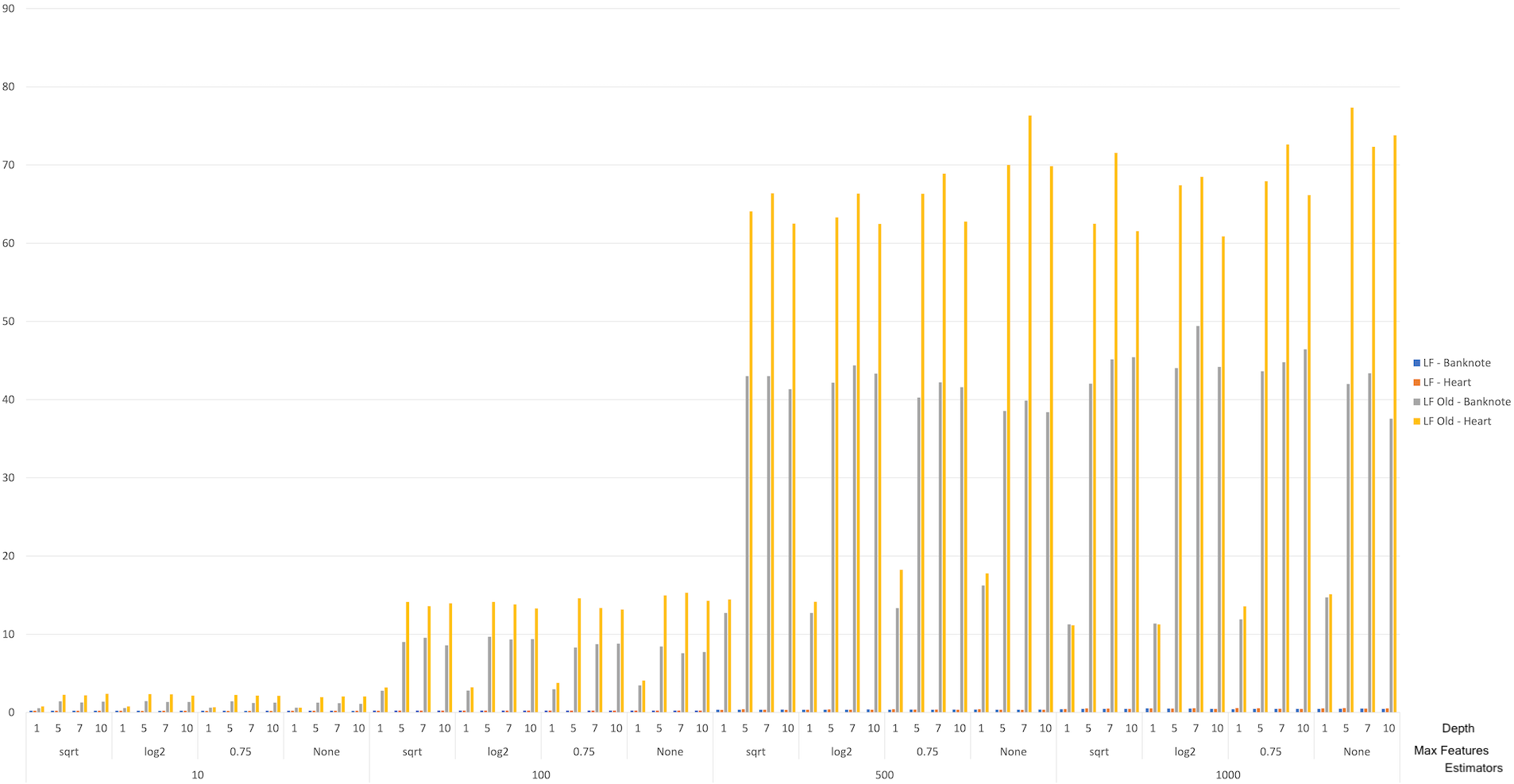}}
\caption{Comparison of preliminary LF and LF on features' ranges generation without reduction (y-axis in seconds)} \label{ta1}
\end{figure}
\begin{figure}[ht]
\centerline{\includegraphics[width=1\textwidth]{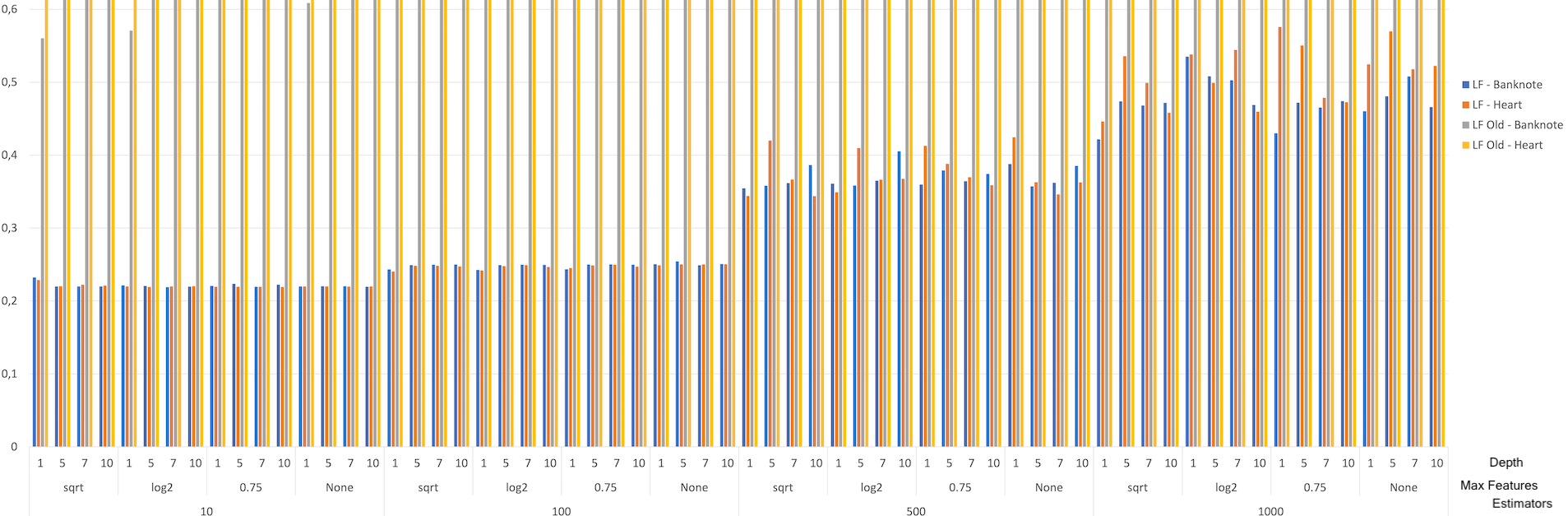}}
\caption{Comparison of preliminary LF and LF on features' ranges generation without reduction (y-axis in second - zoomed)} \label{ta1z}
\end{figure}

The first aspect of the algorithm we are going to inspect, concerns the identification of the ranges for each feature appearing in a rule without applying any reduction method. In Figure~\ref{ta1} and~\ref{ta1z}, we can see that LF is outperforming the preliminary version, with a huge decrease of 93\% for Banknote and 95\% for Heart datasets. In Figure~\ref{ta1z} we are zooming the y-axis in order to make visible that LF runs approximately between 0.2 to 0.6 seconds per explanation, in contrast to the preliminary version which generates explanations from 0.2 to almost 80 seconds.

\begin{figure}[ht]
\centerline{\includegraphics[width=1\textwidth]{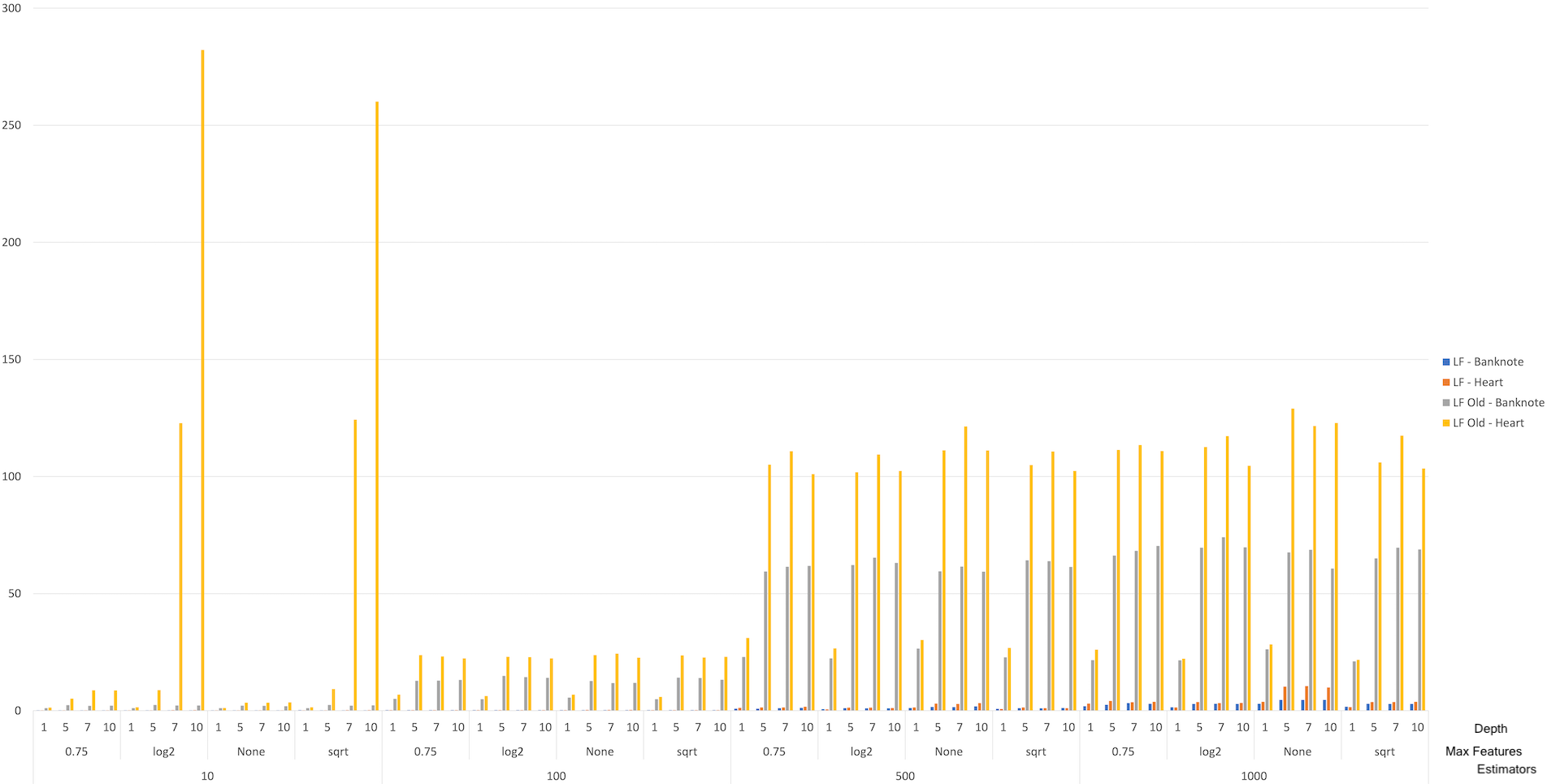}}
\caption{Comparison of preliminary LF and LF on features' ranges generation with reduction (y-axis in second)} \label{ta2}
\end{figure}
\begin{figure}[ht]
\centerline{\includegraphics[width=1\textwidth]{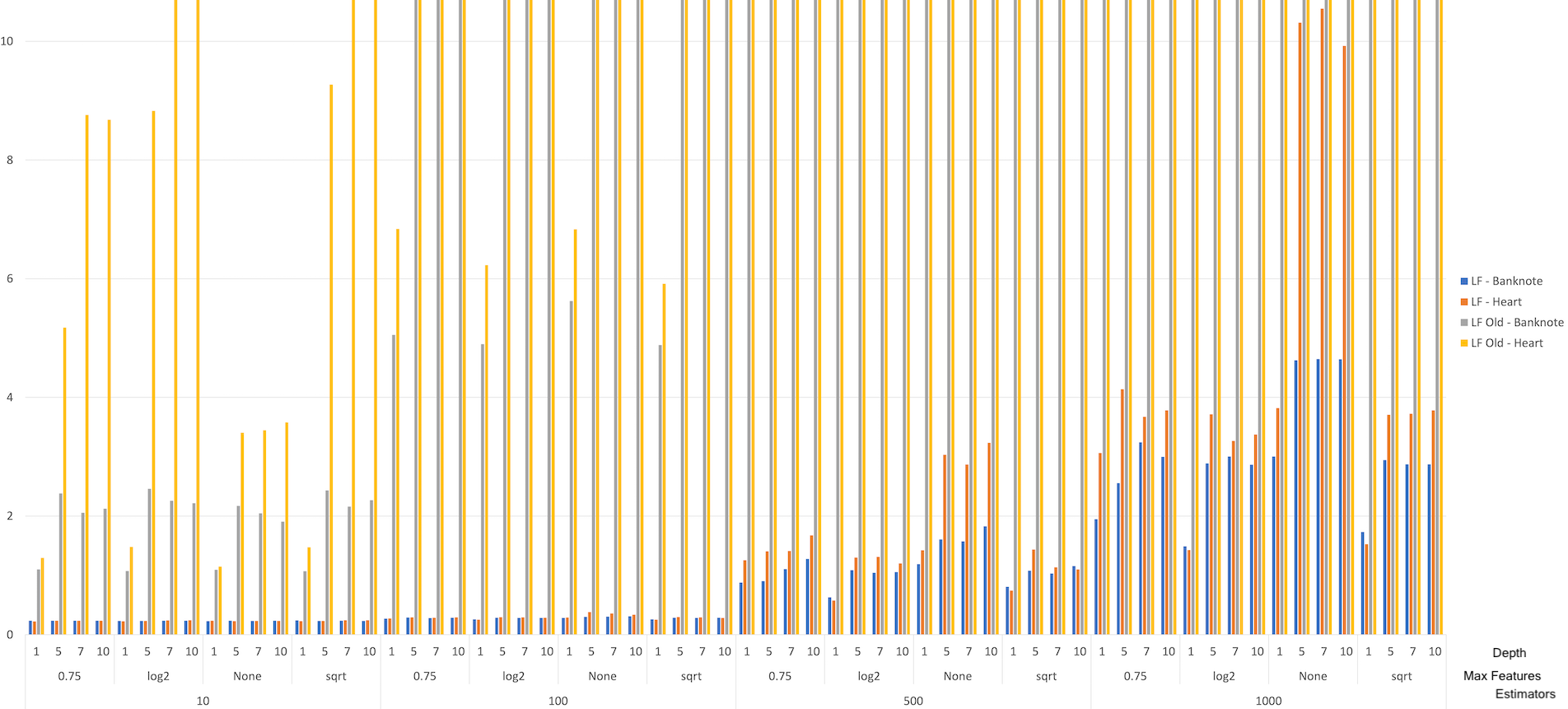}}
\caption{Comparison of preliminary LF and LF on features' ranges generation with reduction (y-axis in second - zoomed)} \label{ta2z}
\end{figure}

Then, we compare the two versions on the generation of rules applying reduction. We remind here that in the reduction pipeline we optimised the reduction through association rules, thus this experiment will inspect the effect of this optimisation. In Figure~\ref{ta2} and~\ref{ta2z}, we can see that LF is once again outperforming the preliminary version, with a huge decrease of 94\% for Banknote and 96\% for Heart datasets. In Figure~\ref{ta2z} we are zooming the y-axis in order to make visible that the version of LF runs approximately between 0.2 to 11 seconds per explanation, in contrast to the preliminary version which generates explanations from 2 to 128 seconds, and even over 280 in few extreme cases.

\subsection{Comparison study with other techniques}
In our experiments we include three techniques from the literature, Anchors~\cite{anchors} (AN), CHIRPS~\cite{badCHIRPS} (CH) and DefragTrees~\cite{satoshiTrees} (DF), as well as a simple surrogate model based on a decision tree in a global (GS) and a local (LS) fashion, to use it as a baseline. AN is a local, model-agnostic interpretation technique for binary and multi-class classification. CH is a local, model-specific technique for interpreting random forest models for binary and multi-class classification tasks. DF is a global, model-specific technique that interprets tree ensembles on binary and multi-class classification and regression. The applicability of each of these techniques to each dataset is presented in Table~\ref{tab:tocomp}. 

\begin{table}[ht]
\resizebox{\textwidth}{!}{%
\begin{tabular}{c|c|c|c|c|c|c|c|c|c|}
\cline{2-10}
 & Banknote &
  \begin{tabular}[c]{@{}c@{}}Heart\\ (Statlog)\end{tabular} &
  \begin{tabular}[c]{@{}c@{}}Adult\\ (Census)\end{tabular} &
  Glass &
  \begin{tabular}[c]{@{}c@{}}Image\\ Segmentation\end{tabular} &
  Abalone (MC) &
  Abalone (R) &
  Boston &
  \begin{tabular}[c]{@{}c@{}}Wine\\ Quality\end{tabular} \\ \hline
\multicolumn{1}{|c|}{\begin{tabular}[c]{@{}c@{}}Baseline\\ Surrogate\end{tabular}} &
  \checkmark & \checkmark & \checkmark & \checkmark & \checkmark & \checkmark & \checkmark & \checkmark & \checkmark \\ \hline
\multicolumn{1}{|c|}{DefragTrees} &
  \checkmark & \checkmark & \checkmark & \checkmark & \checkmark & \checkmark & \checkmark & \checkmark & \checkmark \\ \hline
\multicolumn{1}{|c|}{Anchors} &
  \checkmark & \checkmark & \checkmark & \checkmark & \checkmark & \checkmark & - & - & - \\ \hline
\multicolumn{1}{|c|}{CHIRPS} &
  \checkmark & \checkmark & \checkmark & \checkmark & \checkmark & \checkmark & - & - & - \\ \hline
\end{tabular}%
}
\caption{Applicability of interpretation techniques to each dataset}
\label{tab:tocomp}
\end{table}

We use three different evaluation metrics. The first one computes the length of an explanation for an instance ($rule\_length$). The second one measures the coverage of a rule, which describes the fraction of rules of a dataset that satisfy the antecedent of that rule ($coverage$). The third one is the precision of a rule, which measures the performance of that rule on the covered instances ($precision$), measuring the actual $precision$ in classification problems, and the \textit{mean absolute error} ($mae$) in regression problems. We use the leave-one-out (LOO) cross-validation method in subsets of 10 random samples including the minimum between 10\% or 100 instances for each dataset, to compute these metrics.

Finally, we propose a novel property called ``conclusiveness''. A rule-based explanation method is defined as ``conclusive'' when the antecedent of a rule - explanation - is a definite proof for the consequent. LionForests by design is ``conclusive'' because each rule contains every necessary feature to cover at least the minimum amount of paths, which will always produce the same outcome. However, this automatically suggests that LionForests' explanations will be more specific, and they will have low coverage. In the following, we experimentally prove, with at least one example per method, that all the other methods are not ``conclusive''.

\subsubsection{Comparison with evaluation metrics}

\begin{figure}[ht]
\centerline{\includegraphics[width=0.9\textwidth]{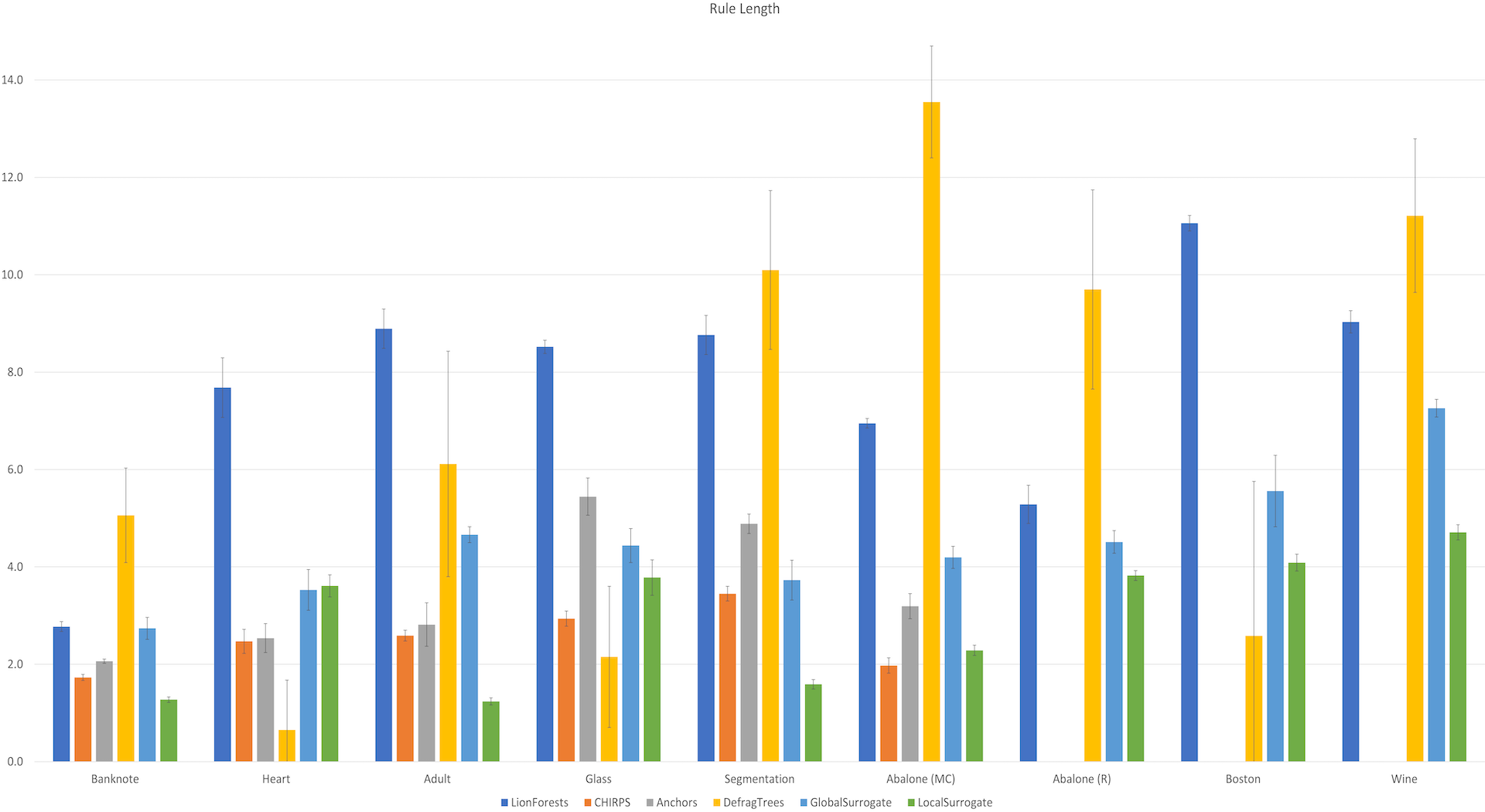}}
\caption{Comparison of techniques in terms of the $rule\_length$ metric} \label{cm1}
\end{figure}

The first metric to be examined is the $rule\_length$. The results of the comparison are visible in Figure~\ref{cm1}. In these results, it is clear that LF and DF are in almost every case the algorithms providing rules with a lot of conditions. Specifically, we can see that LF and DF are producing rules of around $7.66^{\pm0.28}$ and $6.79^{\pm1.7}$ length by average. The other techniques, CH$= 2.53^{\pm0.15}$, AN$= 3.49^{\pm0.27}$, GS$= 4.51^{\pm0.33}$ and LS$= 2.93^{\pm0.15}$, seem to provide smaller rules. CH and AN are not present in the last three columns because these algorithms are not applicable in regression tasks. 

\begin{figure}[ht]
\centerline{\includegraphics[width=0.9\textwidth]{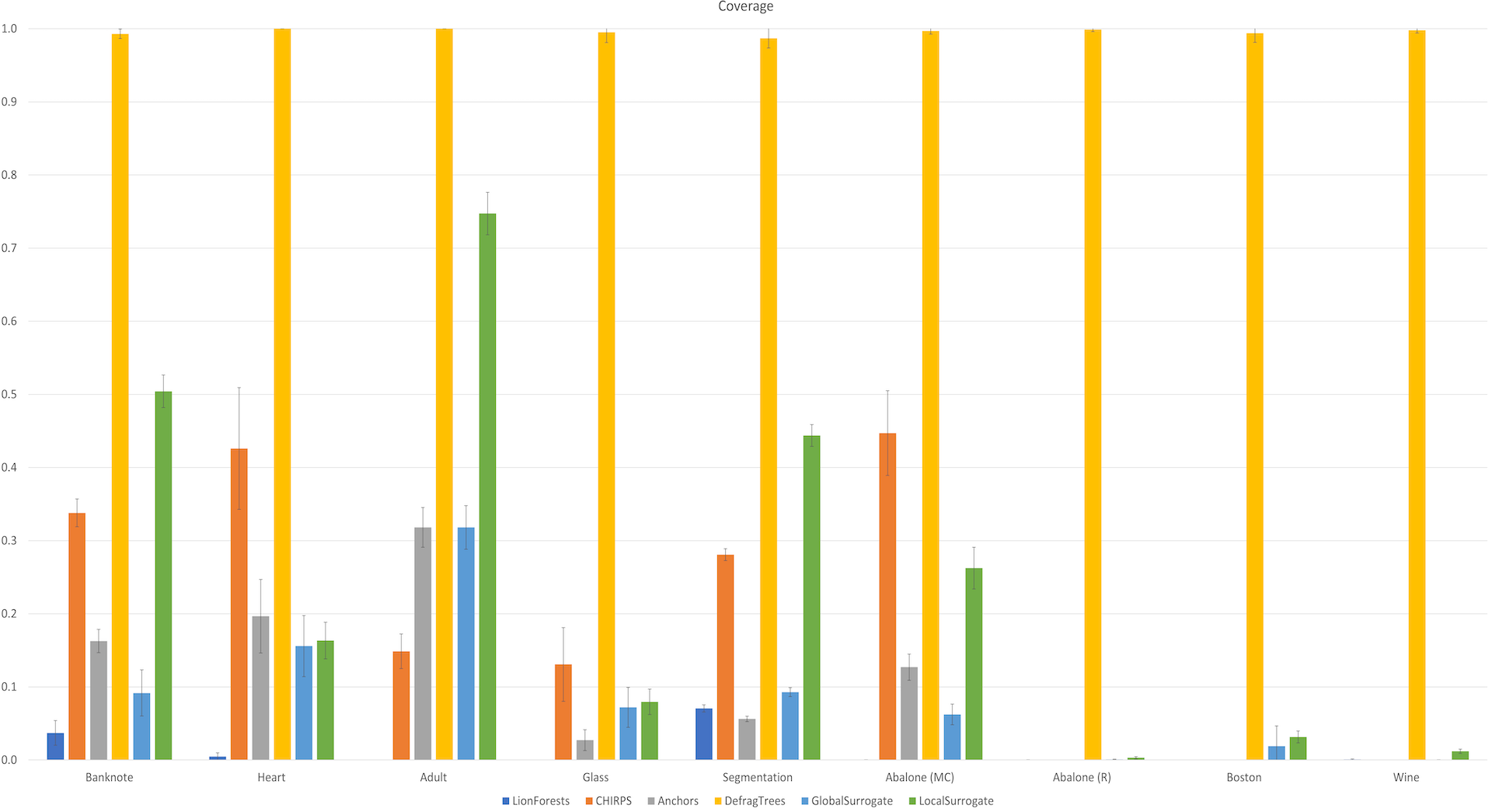}}
\caption{Comparison of techniques in terms of the $coverage$ metric} \label{cm2}
\end{figure}

$Coverage$ is the second metric we investigate (Figure~\ref{cm2}). The most interesting findings in this experiment are the extremely high $coverage$ of DF ($99.59\%^{\pm0.65}$), and the low coverage of LF ($1.26\%^{\pm0.31}$). The performance of the remaining algorithms performed is much lower than the DF's performance, but higher than the LF's performance, CH$= 29.52\%^{\pm4.04}$, AN$= 14.79\%^{\pm2.15}$, GS$= 9.02\%^{\pm1.99}$ and LS$= 24.96\%^{\pm1.67}$.

\begin{figure}[ht]
\centerline{\includegraphics[width=0.9\textwidth]{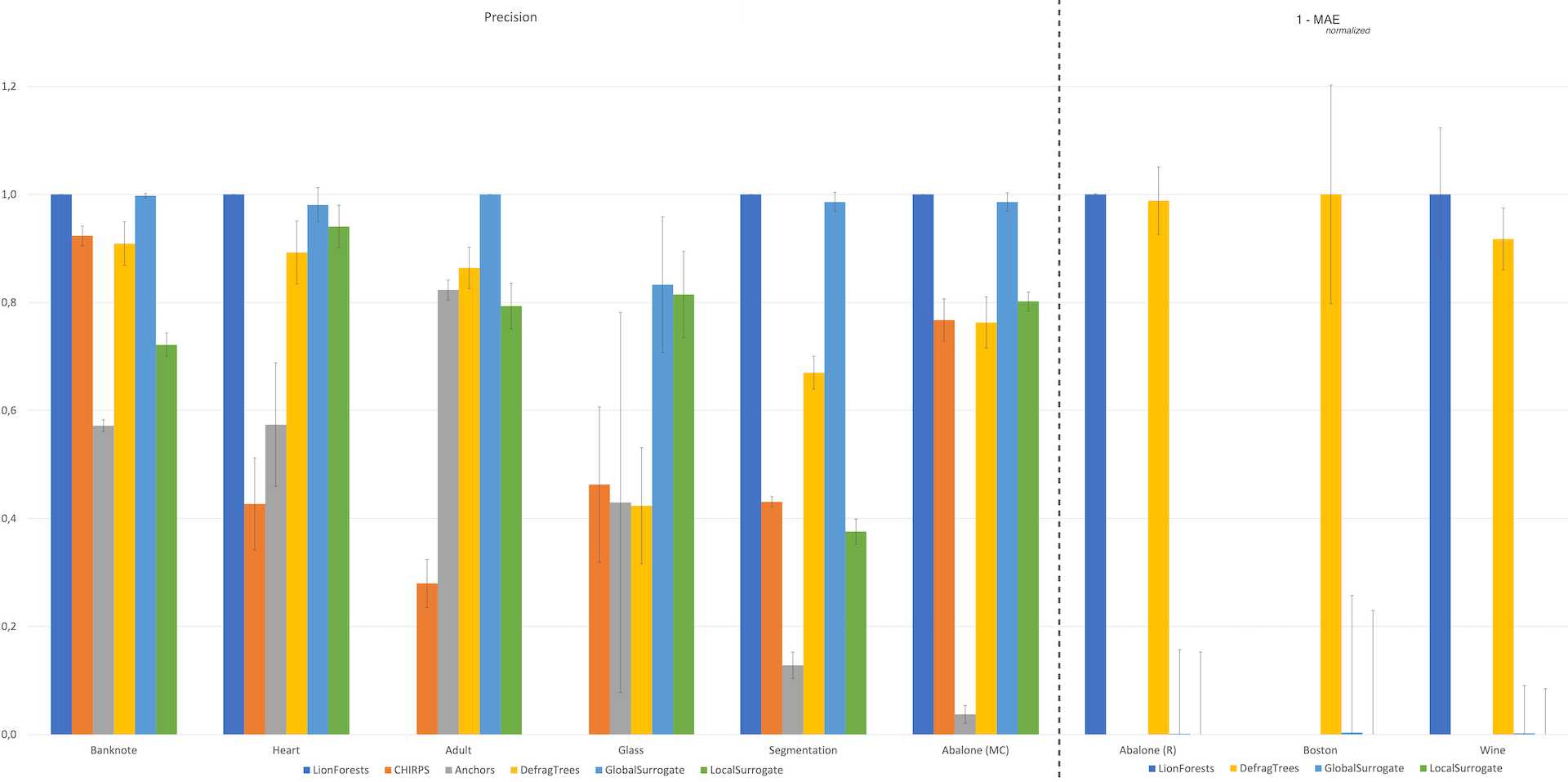}}
\caption{Comparison of techniques in terms of the $precision$ metric and $1-mae_{normalized}$} \label{cm3}
\end{figure}

Finally, we evaluated the techniques using the $precision$ metric. As already mentioned, $precision$ is measured differently across the classification and regression tasks. Therefore, in Figure~\ref{cm3} the first 6 columns are measured using the actual $precision$ of classification tasks, while in regression we used $mae$. Moreover, we scaled the $mae$ error in each of the results of the regression datasets to $[0,1]$ and subtracted by 1 ($1 - mae_{normalised}$) for better comparison and visualisation. The very specific rules of LF achieved a high $100\%^{\pm0.0}$ $precision$ and the highest $1^{\pm0.01}$ $mae$ in all tasks, except from Adult, Glass and Boston datasets, where the coverage was $0$. In the classification tasks, GS had a high $precision$ performance ($96.39\%^{\pm3.28}$), while DF ($75.37\%^{\pm5.38}$) and LS ($74.13\%^{\pm3.74}$) achieved similar to each other results. CH ($54.86\%^{\pm5.67}$) and AN ($42.74\%^{\pm8.94}$) had the worst performance. For the regression tasks, DF resulted in a high score ($0.97\%^{\pm0.11}$), while the performance of GS ($0\%^{\pm0.17}$) and LS ($1\%^{\pm0.16}$) was very low.

\subsubsection{Conclusiveness property investigation}

Even if LF does not always have the best performance in terms of the aforementioned metrics, it has the ability to provide ``conclusive'' interpretation rules, making it trustworthy and competitive. This occurs because LF always retains at least the number of paths that will always produce the same result for a particular instance, and every feature (and its corresponding range) from those paths is included in the final rule (see Section~\ref{sec:concl} and~\ref{mnpr}). In fact, in this section, we demonstrate that all of the algorithms under consideration are not conclusive, by providing at least one example for each one of them. 

We will use the Banknote dataset for this qualitative assessment. We retain 10 of the 1372 instances for manual inspection, and the rest are used to train the RF model. After normalisation, the distributions of Banknote's 4 features $F=[Variance,$ $Skew,Curtosis, Entropy]$ are between $[-1,1]$. Then, we choose a random instance among the 10, $x_1=[0.53, -0.25, -0.24, 0.53]$, which was predicted to be a `fake banknote'. The following rules were provided by the examined interpretation techniques:

\begin{itemize}
    \item [\textbf{LF}:] If $0.36 \leq Variance \leq 1$ and $-0.62 \leq Curtosis \leq 1$ then `fake banknote'
    \item [\textbf{CH}:] If $ 0.23 < Variance$ then `fake banknote'
    \item [\textbf{AN}:] If $ 0.42 < Variance$ then `fake banknote'
    \item [\textbf{LS}:] If $\{\}$ then `fake banknote'
\end{itemize}

LF indicates that the value of the feature $Curtosis$ must be within the range $[-0.62, 1]$, while CH, AN, and LS provide no such requirement. When the instance's value for this feature is set to $-1$, the prediction is changed from ``false banknote'' to ``real banknote''. As a result, the three rules, with the exception of LF, were all not conclusive. We proceed looking at another instance $x_2=[ 0.52, 0.79, -0.91, -0.35]$. We produced the interpretation rules again after RF predicted the instance as ``false banknote''.

\begin{itemize}
    \item [\textbf{LF}:] If $0.43 \leq Variance \leq 1$ and $-0.92 \leq Curtosis \leq -0.89$ then `fake banknote'
    \item [\textbf{DF}:] If $0.34 < Skew$ and $-0.97 < Curtosis$ then `fake banknote'
\end{itemize}

However, we can see that DF lacks a condition for $Variance$, whereas LF lacks a condition for $Skew$. We changed both of them, and LF's decision not to include $Skew$ was correct because there was no change in the outcome when we altered $Skew$ from $0.79$ to either $1$ or $-1$. The prediction, on the other hand, changed when we changed $Variance$ from $0.52$ to $-1$. As a result, DF is not conclusive. Finally, we examined another instance $x_3 = [-0.23,  0.24, -0.73,  0.2]$ to argue that GS is also not conclusive. The RF classified this instance as a ``real banknote''. We obtained the following two interpretation rules after requesting LF and GS to produce them:

\begin{itemize}
    \item [\textbf{LF}:] If $-0.23 \leq Variance \leq -0.12$ and $-1 \leq Skew \leq 0.32$ and $-1 \leq Curtosis \leq -0.57$
    then `real banknote'
    \item [\textbf{GS}:] If $-0.26 < Variance \leq 0.06$ and $0.23 < Skew \leq 0.32$ then `real banknote'
\end{itemize}

It is clear that the rules are not identical. GS omitted the $Curtosis$ feature, while LF provides a condition about this feature. We proceeded to modify this feature's values from $-0.23$ to $1$ and the prediction changed to `fake banknote'. Hence, we can say that GS is also not conclusive.


\subsection{Qualitative evaluation}
\label{sec:visexample}

In this final section of the experiments, we will present an example using an instance from the Adult dataset. Apart from the prediction and the interpretation rule, we will show the interpretation visually, and then perform a few manual tests to guarantee that the interpretation rule is conclusive. Initially, we obtain the following prediction and interpretation rule, with and without LF's reduction, for a random instance of Adult:

\begin{itemize}
    \item [\textbf{Original}:] if \textit{Marital Status} $=$ \textit{Married} and $5119.0\leq$ \textit{Capital Gain}$ \leq 5316.5$ and $0\leq$ \textit{Capital Loss}$ \leq 1782.5$ and $33.5\leq$ \textit{Age}$ \leq 49.5$ and $38.5\leq$ \textit{Hours Per Week}$ \leq 40.5$ and \textit{Occupation} $=$ \textit{Exec Managerial} and $108326.999\leq$ \textit{fnlwgt}$ \leq 379670.501$ and \textit{Education} $=$ \textit{Bachelors} and \textit{Sex} $=$ \textit{Male} and \textit{Workclass} $=$ \textit{Private} and \textit{Native Country} $=$ \textit{United Stated} then $Income > 50K$
    \item [\textbf{LF}:] if \textit{Marital Status} $=$ \textit{Married} and $5119.0\leq$ \textit{Capital Gain}$ \leq 5316.5$ and $31.5\leq$ \textit{Age}$ \leq 49.5$ and $108326.999\leq$ \textit{fnlwgt}$ \leq 379670.501$ then $Income > 50K$
\end{itemize}

Based on the two explanations, we can draw the following conclusions. The capacity of LF to reduce the rules is obvious. The new interpretation is 63\% shorter since it includes only four conditions rather than eleven. The ranges of the features \textit{Age} and \textit{fnlwgt} are also broader in the LF's rule, thus the rule is less sensitive to fluctuations. Finally, we can see that LF reduced nearly all of the categorical features, which is very positive.

\begin{figure}[ht]
\centerline{\includegraphics[width=0.9\textwidth]{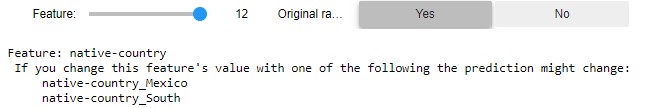}}
\caption{Visualisation of interpretation for feature \textit{Native Country}} \label{ex:ex1}
\end{figure}

We will analyse the features \textit{United States} and \textit{Age}, which are categorical and numerical, respectively, using the visualisation offered by LF. Someone may notice that the feature \textit{United States} does not appear in the particular interpretation rule. This means that the value for \textit{United States} in this case, \textit{United States}, has no effect on the prediction and has been diminished in this instance's interpretation. However, as the visualisation tool in Figure~\ref{ex:ex1} indicates, the only potential values that can influence the prediction are \textit{Mexico} and \textit{South}. 

\begin{figure}[ht]
\centerline{\includegraphics[width=0.8\textwidth]{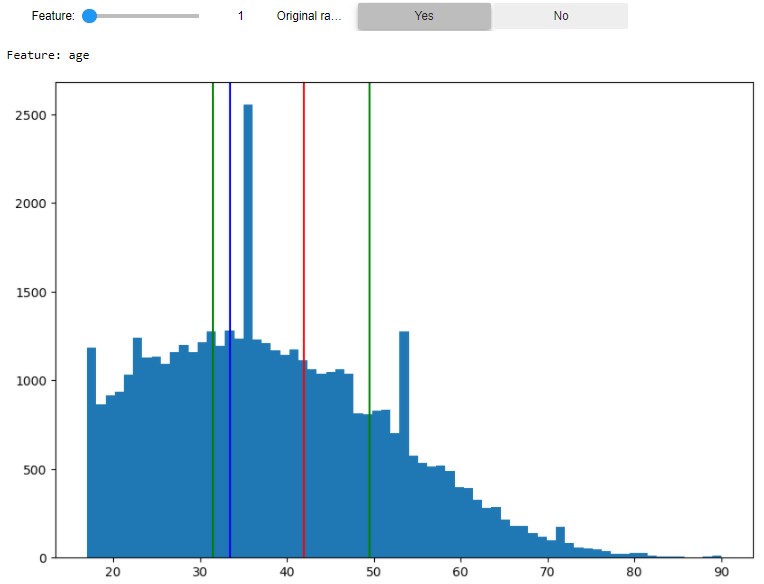}}
\caption{Visualisation of interpretation for feature \textit{Age}} \label{ex:ex2}
\end{figure}

Then, we changed the value of \textit{Native Country} to \textit{Greece}, \textit{United Kingdom}, \textit{Japan} and another 36 countries not appearing in the list of values which may affect the prediction, and the outcome indeed remained the same. Our next modification concerns the \textit{Age} feature, as presented in Figure~\ref{ex:ex2}. Since \textit{Age} is a numerical feature, the distribution of this feature among the training data, as well as the range of the allowed values of this feature for the examined instance are presented in a plot.

This plot firstly informs us about how the ranges of the feature's value were extended. The red vertical line reflects the instance's value for this feature ($41$), while the blue and green lines reflect the initial rule's and reduced rule's ranges, respectively (blue and green right lines are overlapping). We then considered modifying the \textit{Age} with two values within the range, $32$ and $48$, and indeed the classification result remained the same as indicated by the extracted rule.

\section{Discussion}

We have used 8 different datasets from 3 different learning tasks in the above series of experiments to evaluate different aspects of LF and compare it to other techniques.

Via a sensitivity study, we initially investigated how the LF and RF parameters affect the reduction of features and paths (Section~\ref{sec:sensanalysis}). The parameters \textit{`max features'} and \textit{`estimators'} of the RF appear to be strongly associated with the reduction effect. Specifically, when estimators were 500 or greater, the feature reduction ration was always higher.

For the LF's parameters, we discovered that reduction through association rules is required for high feature reduction in all classification scenarios, while reduction through random selection is always required for path reduction. Reduction through clustering seemed to assist in increasing feature reduction, but it was not as effective as reduction through association rules. 

We can notice from the sensitivity analysis of the regression tasks that DSi and DSo are the most promising reduction techniques. We also investigated the impact of the $local\_error$ parameter, which confirmed the hypothesis that higher $local\_error$ would lead to higher feature and path reduction.
 
After the sensitivity analysis, we proceeded to the time analysis in Section~\ref{sec:timeanalysis}. Those experiments proved how much the LF response time was improved in comparison to the preliminary version. Two core procedures were optimised, and the result was a 93\% to 95\% faster run-time on the first procedure, and a 94\% to 96\% on the second. This improvement renders the LF applicable even in online scenarios where immediate interpretations must be provided to users in real-time.

We then compared LF to SOTA algorithms using well-known metrics and a custom property. In terms of $rule\_length$ and $coverage$, LF did not outperform the other techniques, but it did achieve perfect $precision$ in the classification experiments. The low performance of LF with respect to $coverage$ is due to the extremely specific rules offered by LF. At the same time, the rules' specificity renders them larger than the rules of the opposing algorithms. However, LF has the conclusiveness property, which renders it reliable and trustworthy, and by examining the other algorithms we proved that none of them has this property. Thus, despite their superior performance in the metrics favoured in the literature, they lack the ``conclusiveness'' property, which we believe is necessary for any interpretation algorithm that provides rules to the user towards more transparent, complete, and reliable explanations.

We finally presented a detailed use case from one dataset. We chose the Adult dataset which facilitates both numerical and categorical features. By taking a random instance we generated both the prediction by the RF model, the interpretation rule from the LF and the visual interface as well. We investigated both the ranges of a numerical value and the alternate values of a categorical feature using the user interface. In comparison to other algorithms that only provide the feature and the category in the rule, this tool allows the user to see alternate categorical values.

\section{Conclusion}
Random forest is one of the top performing machine learning algorithms in critical sectors such as health, industry, or retail. In the same time, its uninterpretable nature makes it an inappropriate solution, due to trustworthiness concerns, and even issues related to legal frameworks. Therefore, the need of injecting interpretability to such algorithms is evident. In a preliminary work, we introduced a random forest-specific local-based interpretation technique called LionForests. The interpretations are presented in the form of rules. Each rule is a conclusive set of conditions about the features that affected an instance's prediction. LionForests implements a series of feature and path reduction approaches in order to provide smaller rules containing conditions with broader ranges. We are refining this methodology in a number of ways in this work. Providing a stable theoretic background, enhancing its core procedures towards timely responses, extending its applicability to a variety of learning tasks, LionForests technique undergoes a series of experiments. We investigated and evaluated how the parameters of a random forest model, as well as the parameters of LionForests, influence feature and path reduction in these experiments, which were focused on 8 separate datasets. A time analysis backs up our claim that LionForests responds in a timely manner, in contrast to its preliminary version. We compare LionForests using well-known metrics with cutting-edge techniques. However, while the performance of LionForests based on such metrics does not imply that our system is superior, its inherent property known as ``conclusiveness'' distinguishes it from the other techniques. We also demonstrate that all the other techniques are not conclusive. Finally, we give a detailed example of how to best exploit the LionForests technique, in a qualitative manner. Few of our next steps will be to evaluate LionForests in a human-oriented way, exploiting information from relevant works~\cite{VANDERWAA2021103404}, to investigate its applicability to different types of data, as well as to examine different ensemble algorithms.

\section*{Acknowledgements}
This paper is supported by the European Union's Horizon 2020 research and innovation programme under grant agreement No 825619. AI4EU Project\footnote{\url{https://www.ai4eu.eu}}.

%
%

\bibliographystyle{unsrt}  

\end{document}